\documentclass{article} 
\usepackage{iclr2027_conference,times}

\usepackage{amsmath,amsfonts,bm}

\def\eqref#1{equation~\ref{#1}}

\def\1{\bm{1}}

\DeclareMathAlphabet{\mathsfit}{\encodingdefault}{\sfdefault}{m}{sl}
\SetMathAlphabet{\mathsfit}{bold}{\encodingdefault}{\sfdefault}{bx}{n}

\usepackage[utf8]{inputenc} 
\usepackage[T1]{fontenc}    
\usepackage{microtype}      
\usepackage{inconsolata}    
\usepackage{xspace}

\usepackage{hyperref}
\usepackage{cleveref}
\usepackage{url}

\usepackage{amsmath}
\usepackage{amsfonts}
\usepackage{nicefrac}

\usepackage[table]{xcolor}
\usepackage{booktabs}
\usepackage{threeparttable}
\usepackage{multirow}
\usepackage{array}
\usepackage[table]{xcolor}
\usepackage{colortbl}
\usepackage{siunitx}
\usepackage{needspace}
\usepackage{graphicx}
\usepackage{wrapfig}
\usepackage{ragged2e}
\usepackage{subcaption}
\usepackage{placeins}
\usepackage{fontawesome}
\usepackage{float}
\usepackage{tabularx}
\newcolumntype{Y}{>{\RaggedRight\arraybackslash}X}
\usepackage{tabularx}
\usepackage[table]{xcolor}
\usepackage[most]{tcolorbox}
\usepackage{enumitem}
\usepackage{soul}

\definecolor{cellcolor}{RGB}{0,0,0}
\definecolor{antiquewhite}{rgb}{0.98, 0.92, 0.84}
\definecolor{anti-flashwhite}{rgb}{0.95, 0.95, 0.96}
\definecolor{aliceblue}{rgb}{0.94, 0.97, 1.0}
\definecolor{almond}{rgb}{0.94, 0.87, 0.8}
\definecolor{cosmiclatte}{rgb}{1.0, 0.97, 0.91}
\definecolor{darkbyzantium}{rgb}{0.36, 0.22, 0.33}
\definecolor{darkseagreen}{rgb}{0.56, 0.74, 0.56}
\definecolor{darkspringgreen}{rgb}{0.09, 0.45, 0.27}
\definecolor{asparagus}{rgb}{0.53, 0.66, 0.42}
\definecolor{antiquefuchsia}{rgb}{0.57, 0.36, 0.51}
\definecolor{ao(english)}{rgb}{0.0, 0.5, 0.0}
\definecolor{deepcerise}{rgb}{0.85, 0.2, 0.53}
\definecolor{denim}{rgb}{0.08, 0.38, 0.74}
\definecolor{crimson}{rgb}{0.86, 0.08, 0.24}
\definecolor{buff}{rgb}{0.94, 0.86, 0.51}
\definecolor{amber(sae/ece)}{rgb}{1.0, 0.49, 0.0}
\definecolor{airforceblue}{rgb}{0.36, 0.54, 0.66}
\definecolor{amethyst}{rgb}{0.6, 0.4, 0.8}
\definecolor{azure(colorwheel)}{rgb}{0.0, 0.5, 1.0}
\definecolor{azure(web)(azuremist)}{rgb}{0.94, 1.0, 1.0}
\definecolor{beige}{rgb}{0.96, 0.96, 0.86}
\definecolor{cornsilk}{rgb}{1.0, 0.97, 0.86}
\definecolor{darkcerulean}{rgb}{0.03, 0.27, 0.49}
\definecolor{babyblue}{rgb}{0.54, 0.81, 0.94}
\definecolor{ballblue}{rgb}{0.13, 0.67, 0.8}
\definecolor{dukeblue}{rgb}{0.0, 0.0, 0.61}
\definecolor{champagne}{rgb}{0.97, 0.91, 0.81}
\definecolor{antiquefuchsia}{rgb}{0.57, 0.36, 0.51}
\definecolor{darkgreen}{rgb}{0, 0.5, 0}

\definecolor{caseblue}{RGB}{95, 125, 150}
\definecolor{casebluebg}{RGB}{238, 243, 247}

\definecolor{cellshade}{HTML}{F0EDFF}

\newtcolorbox{casebox}[2][]{
    enhanced,
    breakable,
    colframe=caseblue,
    colback=white,
    coltitle=black,
    fonttitle=\bfseries\small,
    fontupper=\small,
    title={#2},
    attach boxed title to top left={xshift=2mm, yshift*=-3mm},
    boxed title style={
        colback=casebluebg,
        sharp corners,
        frame hidden,
    },
    sharp corners,
    boxrule=0.4mm,
    top=-2.5pt,
    bottom=1.5pt,
    left=2.5pt,
    right=2.5pt,
    width=\linewidth,
    #1
}

\newtcolorbox{darkbluecolorbox}{
  colback=darkcerulean!5,
  colframe=darkcerulean!100,
  coltitle=black,
  fonttitle=\bfseries,
  boxrule=1.5pt,
  arc=3pt,
  left=2pt, right=2pt, top=1pt, bottom=1pt
}

\newtcolorbox{greencolorbox}{
  colback=green!3,  
  colframe=green!70!black, 
  coltitle=black,        
  fonttitle=\bfseries,   
  boxrule=1.5pt,         
  arc=3pt,               
  left=2pt, right=2pt, top=1pt, bottom=1pt 
}

\usepackage{algorithm}
\usepackage{algpseudocode}
\usepackage{pifont}
\usepackage{tikz}

\definecolor{headgray}{RGB}{245,245,248}
\definecolor{groupA}{RGB}{232,246,255}
\definecolor{groupB}{RGB}{243,232,255}
\definecolor{groupC}{RGB}{255,240,245}
\definecolor{row1}{RGB}{255,255,255}
\definecolor{row2}{RGB}{250,250,250}

\definecolor{goodgreen}{RGB}{0,128,0}
\definecolor{badred}{RGB}{200,0,0}

\definecolor{almond}{rgb}{0.94,0.87,0.8}
\definecolor{palesilver}{rgb}{0.79,0.75,0.73}
\definecolor{ctrlbg}{RGB}{230,230,250} 
\definecolor{ctrlfg}{RGB}{0,0,0} 
\definecolor{flipbad}{RGB}{255,200,200} 
\definecolor{flipgood}{RGB}{200,255,200} 
\definecolor{shiftblue}{RGB}{55,95,145}

\definecolor{rqblue}{RGB}{35,112,170}
\definecolor{rqlight}{RGB}{238,247,253}

\definecolor{memblue}{RGB}{238,246,252}
\definecolor{gapred}{RGB}{175,45,45}
\definecolor{gapgray}{RGB}{95,95,95}

\definecolor{flipbad}{RGB}{252,235,235}   
\definecolor{flipgood}{RGB}{232,245,242} 

\newcommand{\system}{\textsc{MemTrim}\xspace}

\usepackage{pifont}

\newcommand{\cmark}{\ding{51}\xspace}
\newcommand{\xmark}{\ding{55}\xspace}

\newcommand{\accstd}[2]{\ensuremath{#1_{\scriptstyle\pm #2}}}

\tcbset{
  polarbox/.style={
    breakable,
    enhanced,
    colback=white,
    colframe=black!35,
    boxrule=0.4pt,
    sharp corners,
    arc=2pt,
    titlerule=0pt,
    title style={font=\bfseries}
  }
}

\newtcolorbox{promptbox}[1]{
  polarbox,
  title={#1}
}

\title{
Understanding and Mitigating Inference-Time Overreliance Using Agentic Memory
}

\author{
\textbf{Luoxi Tang}$^{1}$,
\textbf{Yuqiao Meng}$^{1}$,
\textbf{Nilesh Auradkar}$^{1}$,
\textbf{Muchao Ye}$^{2}$,
\textbf{Dazheng Zhang}$^{3}$,
\textbf{Zhaohan Xi}$^{1}$ \\[2mm]
$^{1}$Binghamton University, State University of New York,
Binghamton, NY, USA \\
$^{2}$University of Iowa,
Iowa City, IA, USA \\
$^{3}$Penn Medicine, University of Pennsylvania,
Philadelphia, PA, USA
}

\iclrfinalcopy

\AddToHook{cmd/maketitle/after}{
    \fancyhf{}
    \fancyhead[L]{Preprint.}
    \renewcommand{\headrulewidth}{0.4pt}
    \pagestyle{fancy}
    \thispagestyle{fancy}
}

\begin{document}

\maketitle


\begin{abstract}
Agentic memory allows LLM agents to reuse past experience, yet retrieved
memories can also distort inference even when they are benign, correctly
stored, and appropriately retrieved. We study this failure mode, which we call
\emph{memory overreliance}. Across benchmarks and memory architectures, we
find that memory is useful when past experience transfers to the current task,
but can become misleading when only part of the evidence transfers. Failures
are strongest under partial query-memory overlap, a pattern further confirmed
by controlled experiments that vary the amount of overlapping evidence.
Motivated by this finding, we propose \system, a plug-and-play framework that
indexes memory evidence at write time and controls its reuse at read time.
\system removes repeated or conflicting evidence while preserving useful
memory-specific information, requires no retraining, and applies to both
embedding-based and structured memory systems. Experiments show that \system
reduces memory overreliance while preserving the benefits of useful memory
across models and memory settings.
\end{abstract}

\section{Introduction}
\label{sec:intro}

Memory allows LLM agents to carry information across interactions and reuse
past experience on future tasks. This capability is increasingly important
for agents that operate over long horizons, where solving each request in
isolation would discard useful experience accumulated from previous
interactions~\citep{xiong-etal-2026-memory,park2023generative,
zhong2024memorybank,shinn2023reflexion,zhao2024expel}. However, memory can
misguide inference even when the memory itself is benign. A memory may be
correctly formed and appropriately retrieved during normal user-agent
interactions, yet contain information that does not fully transfer to the
current task~\citep{yoran2024making,yu2024chain,kim2024re}. Once placed in the
inference context, such memory can exert undue influence on the model's
reasoning. We refer to this phenomenon as \textbf{memory overreliance}:
excessive reliance on benign memory that harms inference on the current task.

{\bf Our Work.} We study this issue in three steps. We first establish when
memory overreliance occurs and investigate its cause. We then develop a
mitigation method. Finally, we test whether the method reduces overreliance
while preserving the benefits of useful memory.

We begin with a three-level empirical analysis of memory overreliance.
\ding{182} First, we ask \emph{whether overreliance arises under ordinary
memory use}. Across multiple benchmarks~\citep{cohen2024evaluating,
joishy2025flying,jin2023cladder} and memory
architectures~\citep{chhikara2025mem0,xu2026mem,packer2023memgpt,
rasmussen2025zep}, we find a clear contrast: memory improves inference when
the stored interaction remains applicable, but can reduce accuracy when the
current task shares substantial evidence with the memory while requiring a
different conclusion. \ding{183} Subsequently, we examine \emph{where these failures occur}. We compare
query-memory overlap between successful and failed cases and find a
statistically significant difference. Successful cases appear more often near
the two extremes of the overlap range, while failures concentrate under
partial overlap. In this regime, enough evidence is shared to make the memory
influential, yet some answer-relevant information still differs. \ding{184} Furthermore, we test this relationship directly through controlled experiments that vary the amount of overlapping evidence
while holding the remaining memory content fixed. We observe an effect that, with little overlap, memory has limited influence; as overlap
increases, its negative effect grows and is strongest at intermediate levels.
The effect then reverses as the interaction becomes sufficiently aligned with
the stored experience. Collectively, these results indicate that partial evidence
overlap is a key reason in which ordinary memory use can become
misleading at inference time.

\begin{figure}[t]
    \centering
    \includegraphics[width=0.9\linewidth]{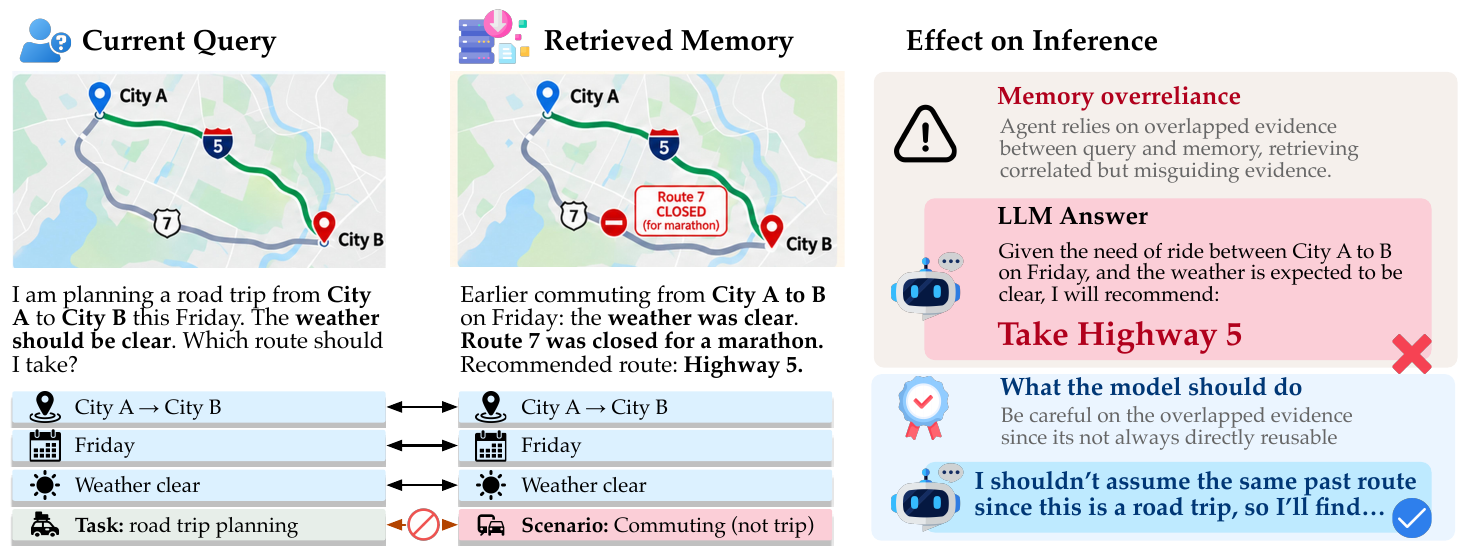}
    \vspace{-3pt}
    \caption{
    Illustration of memory overreliance. A retrieved memory can appear highly
    relevant because it shares substantial evidence with the current query,
    while still carrying context-specific information that does not transfer
    to the present task. Reusing the memory without separating these pieces of
    evidence can misguide the model toward an unsupported conclusion.
    }
    \label{fig:intro}
    \vspace{-5pt}
\end{figure}

These findings motivate \system\footnote{Memory with \emph{Trie-based
Replicate-aware Inference Mitigation}.}, a plug-and-play framework that controls
how overlapping memory evidence enters inference. At write time, \system
decomposes each memory into evidence units and indexes them in a
trie~\citep{bodon2003trie}. Identical evidence maps to shared trie nodes, while
different values of the same evidence branch separately. At read time, the
current query is mapped into the same evidence space and compared with the
retrieved memories. \system removes evidence already present in the query,
drops conflicting memory evidence, and retains useful information found only
in memory. It also removes a previous answer when the evidence it relied on has
changed, or preserves it only as contextual evidence when the current query
asks a different question. Since \system operates only when memory is written
and read, it can be added to existing embedding-based and structured memory
systems without retraining the underlying LLM.

We evaluate whether this design reduces overreliance without weakening useful
memory. Across benchmarks, LLMs, and memory architectures, \system reduces
memory-induced errors on task-divergent cases while preserving the benefit of
memory when the stored interaction remains applicable. Component ablations
show that removing repeated and conflicting evidence, handling previous
answers when relevant evidence changes, and distinguishing task changes each
contribute in the settings they are designed for. We further test \system
under deliberate memory attacks and find that its performance degrades more
gradually as misleading memories increase. Finally, \system introduces only
small changes in inference time across the evaluated memory systems.

To summarize, our contributions are as follows:
\begin{itemize}
    \item We identify \emph{memory overreliance} as an inference-time failure
    of benign agentic memory and show that failures concentrate under partial
    evidence overlap between the current query and retrieved memory.

    \item We propose \system, a plug-and-play framework that indexes memory at
    the evidence level and controls how repeated, conflicting, and
    memory-specific information is reused during inference.

    \item We evaluate \system across benchmarks, LLMs, and memory
    architectures, including component ablations, memory-attack stress tests,
    and runtime analysis.
\end{itemize}
\section{An Empirical Study of Memory Overreliance}
\label{sec:overreliance}

\vspace{-5pt}
We study memory overreliance through three questions. RQ1 asks whether benign
memory can hurt inference under normal memory use. RQ2 examines where these
failures occur along the query-memory overlap range. RQ3 then controls the
amount of overlapping evidence to test whether the same pattern persists when
the overlap is varied directly.

\textbf{Benchmarks and Models.}
We evaluate on RippleEdits~\citep{cohen2024evaluating}, CounterLogic~\citep{joishy2025flying}, and
CLADDER-Derived~\citep{jin2023cladder} using Qwen3.8-27B and GPT-5.6 Luna.
For each benchmark, memories are built from earlier interactions with valid
stored answers and evaluated on separate queries. 

\textbf{Memory systems.}
We evaluate four memory systems with different architectures:
Mem0~\citep{chhikara2025mem0} stores extracted memory entries in a retrieval
index; A-Mem~\citep{xu2026mem} creates structured notes and dynamically links
related memories; MemGPT~\citep{packer2023memgpt} separates active and archival
memory; and Graphiti~\citep{rasmussen2025zep} represents memory as a relational graph.

\textbf{Cases and metrics.}
We divide queries into \emph{task-aligned} (TA) and
\emph{task-divergent with overlapping evidence} (TD-Dup) cases. In TA cases,
the current query asks the same task as the stored interaction and
preserves the answer-relevant evidence, so the retrieved memory is
useful. In TD-Dup cases, the query still shares substantial evidence with the
stored interaction, but the task or answer-relevant information differs (outdated, conflicting, or irrelevant), so
the stored conclusion should not be directly reused. We report task accuracy
for both groups. On TD-Dup cases, we additionally track predictions that
change from correct to wrong
(\cmark$\rightarrow$\xmark) and from wrong to correct
(\xmark$\rightarrow$\cmark) after memory is introduced, separating errors
caused by memory from cases where memory improves the prediction.

\textbf{RQ1: Can ordinary memory use degrade inference?}

We first compare each model with and without memory. Because every stored interaction was correct when written, this
experiment isolates whether ordinary memory reuse can have negative impacts once the
current memory storage no longer supports the same query task.


\definecolor{ctrlfg}{gray}{0.2}
\definecolor{ctrlbg}{gray}{0.96}

\begin{table}[t]
\centering
\scriptsize

\setlength{\tabcolsep}{1.8pt}
\setlength{\aboverulesep}{1pt}
\setlength{\belowrulesep}{1pt}
\renewcommand{\arraystretch}{1.0}

\caption{
Accuracy (\%) on task-aligned (TA) and task-divergent with duplicated evidence
(TD-Dup) cases. TA denotes cases where the memory storage is directly
applicable, while TD-Dup denotes cases with substantial query-memory overlap
but a non-transferable stored conclusion.
\cmark$\rightarrow$\xmark and
\xmark$\rightarrow$\cmark count memory-induced regressions and corrections on
TD-Dup cases, respectively. Standard deviations are shown as subscripts.
}

\vspace{-7pt}

\begin{tabular*}{\textwidth}{
@{\extracolsep{\fill}}
ll
>{\columncolor{ctrlbg}\color{ctrlfg}}c c
>{\columncolor{flipbad}}c
>{\columncolor{flipgood}}c
>{\columncolor{ctrlbg}\color{ctrlfg}}c c
>{\columncolor{flipbad}}c
>{\columncolor{flipgood}}c
>{\columncolor{ctrlbg}\color{ctrlfg}}c c
>{\columncolor{flipbad}}c
>{\columncolor{flipgood}[\tabcolsep][0pt]}c
@{}
}

\toprule

& &
\multicolumn{4}{c}{\textbf{RippleEdits}} &
\multicolumn{4}{c}{\textbf{CounterLogic}} &
\multicolumn{4}{c}{\textbf{CLADDER-Derived}} \\

\cmidrule(lr){3-6}
\cmidrule(lr){7-10}
\cmidrule(lr){11-14}

\textbf{Model}
& \textbf{Memory}
& \textbf{TA}
& \textbf{TD-Dup}
& \ding{51}$\rightarrow$\ding{55}
& \ding{55}$\rightarrow$\ding{51}
& \textbf{TA}
& \textbf{TD-Dup}
& \ding{51}$\rightarrow$\ding{55}
& \ding{55}$\rightarrow$\ding{51}
& \textbf{TA}
& \textbf{TD-Dup}
& \ding{51}$\rightarrow$\ding{55}
& \ding{55}$\rightarrow$\ding{51} \\

\midrule

\multirow{5}{*}{Qwen3.8-27B}
& No Mem.  & \accstd{68.0}{0.4} & \accstd{71.3}{0.8} & --   & --  & \accstd{85.4}{1.3} & \accstd{86.1}{0.9} & --  & --  & \accstd{60.1}{0.5} & \accstd{61.2}{0.9} & --   & --  \\
& Mem0     & \accstd{85.3}{0.7} & \accstd{62.6}{1.3} & 11.6 & 2.9 & \accstd{93.1}{0.9} & \accstd{80.0}{1.4} & 7.8 & 1.7 & \accstd{79.6}{0.8} & \accstd{46.6}{1.5} & 16.9 & 2.3 \\
& A-Mem    & \accstd{82.6}{1.0} & \accstd{61.2}{1.5} & 12.8 & 2.7 & \accstd{92.4}{0.6} & \accstd{80.6}{1.2} & 7.1 & 1.6 & \accstd{78.4}{0.9} & \accstd{46.1}{1.4} & 17.6 & 2.5 \\
& MemGPT   & \accstd{80.9}{0.8} & \accstd{65.7}{1.1} & 8.9  & 3.3 & \accstd{91.2}{1.1} & \accstd{82.8}{1.0} & 5.2 & 1.9 & \accstd{78.1}{0.7} & \accstd{49.8}{1.2} & 14.1 & 2.7 \\
& Graphiti & \accstd{87.4}{0.5} & \accstd{58.3}{1.7} & 15.4 & 2.4 & \accstd{94.6}{0.7} & \accstd{78.7}{1.6} & 8.9 & 1.5 & \accstd{81.2}{0.6} & \accstd{43.9}{1.8} & 19.4 & 2.1 \\

\midrule

\multirow{5}{*}{GPT-5.6 Luna}
& No Mem.  & \accstd{65.4}{0.8} & \accstd{68.9}{0.7} & --   & --  & \accstd{84.9}{1.2} & \accstd{84.7}{0.8} & --  & --  & \accstd{58.8}{0.6} & \accstd{58.4}{0.8} & --   & --  \\
& Mem0     & \accstd{86.1}{0.4} & \accstd{62.2}{1.2} & 9.7  & 3.0 & \accstd{93.4}{0.7} & \accstd{79.4}{1.3} & 7.0 & 1.7 & \accstd{80.4}{0.9} & \accstd{45.5}{1.1} & 15.3 & 2.4 \\
& A-Mem    & \accstd{83.7}{1.0} & \accstd{60.8}{1.1} & 10.9 & 2.8 & \accstd{92.2}{0.9} & \accstd{80.1}{1.0} & 6.4 & 1.8 & \accstd{78.9}{1.1} & \accstd{46.2}{1.6} & 14.8 & 2.6 \\
& MemGPT   & \accstd{82.4}{0.5} & \accstd{65.5}{0.9} & 6.8  & 3.4 & \accstd{91.6}{0.6} & \accstd{82.1}{0.8} & 4.6 & 2.0 & \accstd{79.2}{1.0} & \accstd{48.6}{1.1} & 12.6 & 2.8 \\
& Graphiti & \accstd{88.0}{0.4} & \accstd{57.8}{1.5} & 13.6 & 2.5 & \accstd{94.3}{0.5} & \accstd{77.9}{1.4} & 8.3 & 1.5 & \accstd{82.1}{0.6} & \accstd{43.5}{1.4} & 17.1 & 2.2 \\

\bottomrule

\end{tabular*}

\label{tab:rq1}

\end{table}

\vspace{-5pt}

{\bf Observations:} Table~\ref{tab:rq1} reveals a clear difference between task-aligned (TA) and
task-divergent with overlapping evidence (TD-Dup) cases. In TA cases, retrieved
memory is consistently useful across datasets and memory architectures because
the stored interaction remains applicable to the current task. In TD-Dup
cases, the same memory can instead reduce accuracy. At the example level,
memory-induced regressions occur much more often than memory-induced
corrections.

{\bf Insights:} This contrast suggests that the problem is not that memory is generally noisy
or unhelpful. The same memory mechanism benefits inference when the stored
experience transfers to the current task, yet becomes misguiding when only part
of that experience transfers. Memory overreliance therefore arises from how
partially overlapping evidence is reused during inference.

\textbf{RQ2: Where does memory overreliance occur?}

We next examine which TD-Dup cases are the most vulnerable. We call a case
\emph{successful} when it remains correct after memory is added and
\emph{failed} when memory changes a correct prediction into an incorrect one.
For each case, we compute the content-word Jaccard overlap between the current
query and retrieved memory, excluding the stored answer. We refer to this
quantity as \emph{query-memory overlap} and compare its distribution between
the two groups.

\begin{figure}[t]
\centering
\includegraphics[width=\linewidth]
{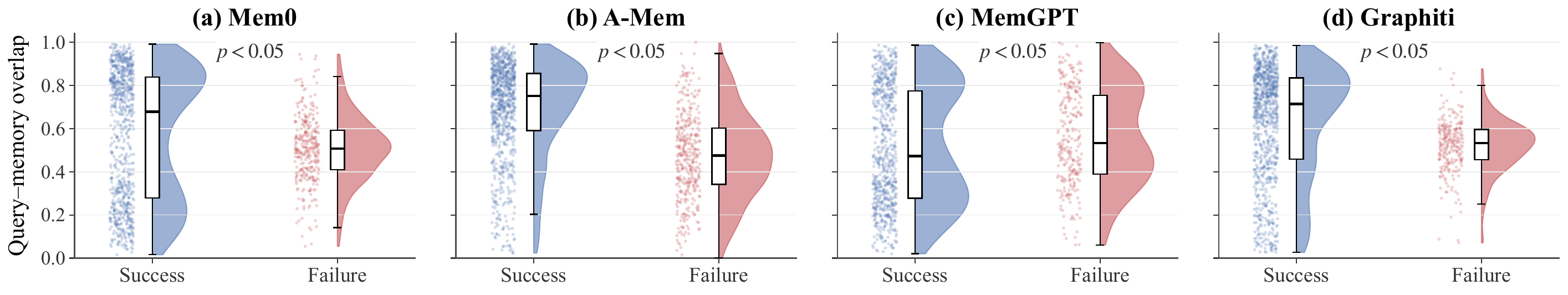}
\captionsetup{font=footnotesize}
\vspace{-5pt}
\caption{
Query-memory overlap (Jaccard similarity) for successful and failed TD-Dup cases on RippleEdits.
The statistical test~\citep{simonsohn2018two} assesses whether the overlap distributions differ between the two groups.
}
\label{fig:rq2}
\end{figure}

{\bf Observations:} Figure~\ref{fig:rq2} shows that successful and failed cases occupy different
parts of the overlap range. Successful cases occur more often near the two
extremes: very little overlap gives memory limited influence, while very high
overlap indicates that the current task closely resembles the stored
interaction. Failed cases instead concentrate under partial overlap, with a
statistically significant separation across all four memory systems.

{\bf Insights:} This pattern suggests that overlap plays two roles. Increasing overlap makes
retrieved memory more influential but does not guaranty that the evidence
supporting its conclusion also transfers to the current task. Partial overlap
is therefore particularly vulnerable: enough evidence is overlapped  to make
the memory dominant, while some answer-relevant information can still diverge.
The exact distribution varies across memory architectures, but this pattern
remains consistent. More results are provided in
Appendix~\ref{app:other-dataset}.

\textbf{RQ3: How does overreliance change with overlapping evidence?}

\begin{figure}[t]
\centering
\includegraphics[width=\linewidth]{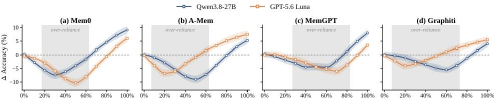}
\captionsetup{font=footnotesize}
\vspace{-5pt}
\caption{
Change in accuracy as the amount of overlapping evidence is varied from
task-divergent toward task-aligned interaction. The y-axis reports the
accuracy change relative to inference without memory. The gray region marks
the range where memory is most misguiding on average.
}
\label{fig:rq3}
\end{figure}

RQ2 examines TD-Dup cases as they naturally occur, where factors beyond overlap
may also distinguish successful from failed cases. To isolate the role of
overlap, we construct a controlled setting in which each memory is
decomposed into evidence units and vary the fraction repeated in the current
query while keeping the remaining content fixed. As overlap increases, the
answer-relevant evidence is gradually restored and the interaction moves toward
the TA setting. At each level, we measure the memory effect as
$\Delta\mathrm{Acc}=
\mathrm{Acc}_{\mathrm{mem}}-\mathrm{Acc}_{\mathrm{no\text{-}mem}}$ and test
the non-monotonic trend with a quadratic term and the two-lines
test~\citep{simonsohn2018two}.

{\bf Observations:} Figure~\ref{fig:rq3} shows a consistent non-monotonic trend across models
and memory systems. With little overlapping evidence, memory has limited
influence. As overlap increases, its harmful effect grows and is strongest
at intermediate overlap. The effect then reverses as the current query becomes
more aligned with the stored interaction.

{\bf Insights:} This observation further explains RQ2. overlap first increases memory influence without
making the stored conclusion valid for the current task. At intermediate
overlap, substantial evidence is repeated while some decisive information
still differs. As the interaction approaches TA, that answer-relevant evidence
is restored and the stored conclusion becomes reusable. Memory overreliance
therefore emerges when evidence overlap is high enough to amplify memory, but
task alignment is still incomplete.

\begin{greencolorbox}
\underline{\textbf{Overall Insight \faLightbulbO.}} RQ1-RQ3 consistently deliver an empirical insight:
memory is useful in task-aligned cases, but can be
misguiding \textbf{when overlapping evidence is substantial but not fully aligned with the task}, which motivates our subsequent mitigation strategy.
\end{greencolorbox}

\section{Method}
\label{sec:method}

\begin{figure}[t]
    \centering
    \includegraphics[width=\linewidth]{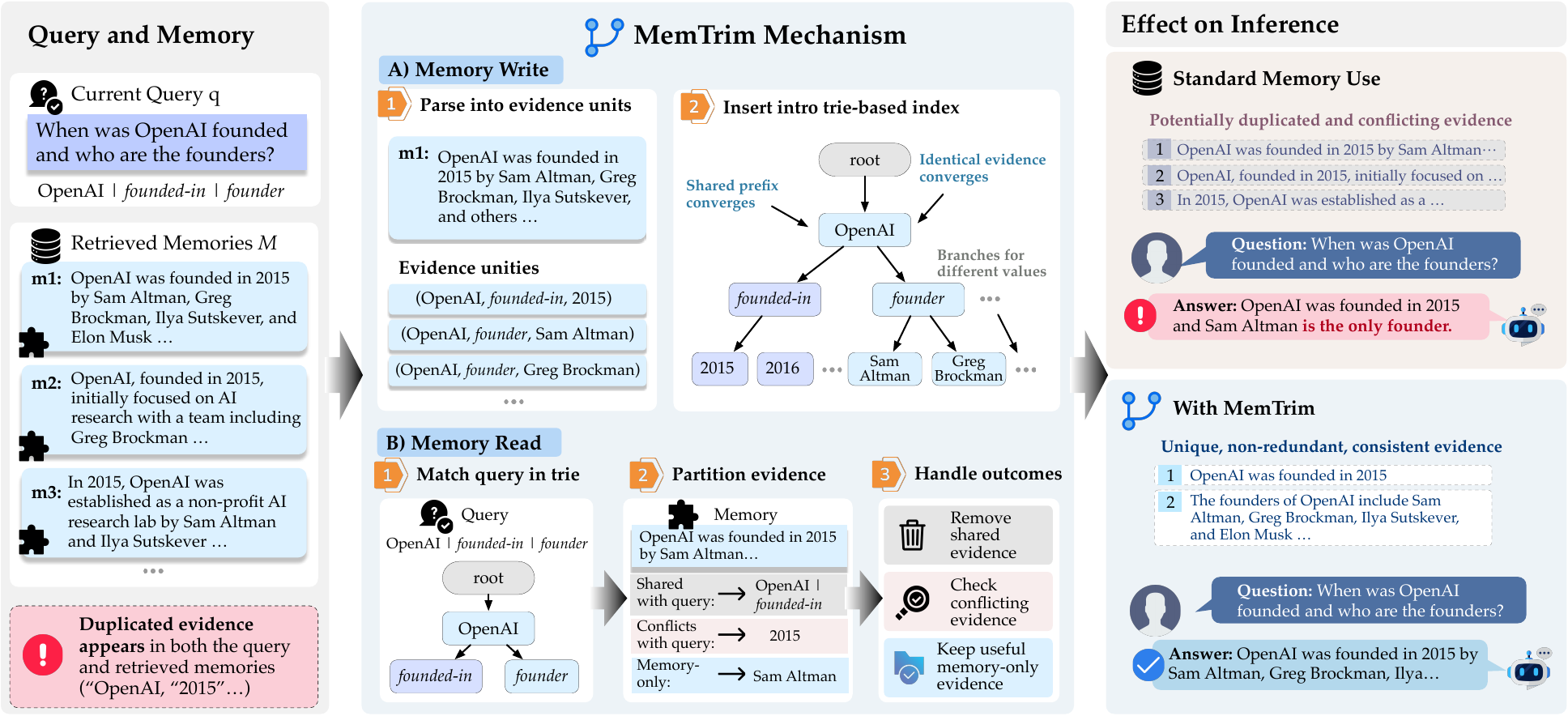}
    \vspace{-7pt}
    \caption{
    Overview of \system. At write time, memories are decomposed into evidence
    units and linked through a trie-backed index. At read time, \system
    identifies shared, conflicting, and memory-only evidence, then constructs
    a non-redundant memory context before inference.
    }
    \label{fig:method}
    \vspace{-10pt}
\end{figure}

Our empirical study shows that memory overreliance is most influential when a
retrieved memory partially overlaps with the current query. This motivates a
simple mitigation principle: \textbf{explicitly identify replicated evidence
and control how it enters inference.} We develop \system to implement this
principle. Figure~\ref{fig:method} gives an overview. We next describe the
evidence representation (\S\ref{ssec:method-overview}), write-time indexing
(\S\ref{ssec:method-write}), and read-time memory use
(\S\ref{ssec:method-read}).

\subsection{Overview and Evidence Representation}
\label{ssec:method-overview}

\system is a plug-and-play layer around an existing memory system. The
underlying system continues to store complete memories and retrieve relevant
entries using its original mechanism. \system adds an evidence index when a
memory is written and processes retrieved memories before they are passed to
the LLM. It therefore requires neither model retraining nor replacement of
the original memory architecture.

Consider a stored interaction $m=(x_m,y_m)$, where $x_m$ contains the query
and its context and $y_m$ is the resulting answer. We parse $x_m$ into a set
of atomic evidence units $E_m$. Each unit is represented as a key-value pair
$e=(k,v)$, where $k$ identifies an attribute or relation and $v$ gives its
value. We further extract a task signature $\tau_m$ that specifies what the
interaction asks the model to determine. The stored answer is represented as
a typed outcome $o_m=(\tau_m,y_m)$ and associated with a support set
$H_m\subseteq E_m$, which contains the evidence on which that outcome depends.
Thus, a memory is represented as
$m=(E_m,\tau_m,H_m,o_m)$. The same parser is applied to a current query $q$,
producing $E_q$ and $\tau_q$. For free-text memory, we obtain these fields with
a fixed instruction-based parser. If the underlying memory already contains
structured metadata (e.g., graph-based memory \citep{rasmussen2025zep}), they
can be mapped directly into the same representation. Details of implementations
(e.g., evidence parsing and alignment) are provided in
Appendix~\ref{app:implementation}.

\begin{casebox}{\bf Illustrative Example}
Consider a stored interaction:
\emph{``Northstar Robotics was founded in 2016 by Maya Lee and Daniel Ortiz.
When was Northstar Robotics founded?''}
The stored answer is \emph{``2016.''}

\system parses the interaction into evidence units such as
$(\texttt{entity},\texttt{NorthstarRobotics})$,
$(\texttt{founded\_in},2016)$,
$(\texttt{founder},\texttt{MayaLee})$, and
$(\texttt{founder},\texttt{DanielOrtiz})$.
Its task signature is $\tau_m=\texttt{founding\_year}$, while the stored
outcome is $o_m=(\texttt{founding\_year},2016)$. Since the answer depends
directly on the founding-year evidence,
$(\texttt{founded\_in},2016)\in H_m$.
\end{casebox}

\subsection{Memory Write: Trie-Backed Evidence Index}
\label{ssec:method-write}

At write time, \system inserts each evidence unit into a global trie
$\mathcal{T}$. Evidence units are indexed independently, so one memory can be
linked to multiple trie nodes. For an evidence unit $e=(k,v)$, the trie first
follows its canonical key $k$ and then its value $v$. The corresponding node
records the memories that contain this evidence and whether the unit belongs
to their support sets $H_m$. Complete memories remain in the original memory
store; \system maintains only this evidence-level index.

The trie makes shared and changed evidence directly distinguishable.
Memories containing the same key-value unit point to the same node, whereas
different values of the same key occupy different branches. For example,
$\texttt{founded\_in=2016}$ and
$\texttt{founded\_in=2017}$ share the same relation key but diverge at the
value level.

{\bf Sparse evidence overlap.}
Shared evidence need not form a continuous prefix between a memory and a
query. Suppose
$E_m=\{e_1,e_2,e_3,e_4\}$ and
$E_q=\{e_1,e'_2,e_3,e'_4\}$.
Because each unit is indexed independently, \system can still match $e_1$ and
$e_3$ even though the intervening units differ. This allows the index to
capture replicated evidence at arbitrary positions within an interaction.

\subsection{Memory Read: Replicate-Aware Context Construction}
\label{ssec:method-read}

At inference time, the underlying memory system first retrieves its usual
candidate set $\mathcal{R}(q)$. \system parses the current query into $E_q$
and looks up its evidence units independently in the trie. For each retrieved
memory $m$, the resulting matches divide its evidence into three sets:
(i) \emph{shared evidence} $S_m(q)$, whose key and value also appear in the
query; (ii) \emph{conflicting evidence} $D_m(q)$, whose key appears in the
query with a different value; and (iii) \emph{memory-only evidence} $U_m(q)$,
which does not appear in the query. Since matching is performed independently
for each unit, shared evidence can occur anywhere in the memory.

\system then constructs the memory context given to the LLM. Shared evidence
is removed from the retrieved memory because it is already stated in the
query, avoiding repeated support for the same information. Conflicting
evidence is also removed because the current query supplies the value for the
current inference. Memory-only evidence is retained when it provides
additional information absent from the query.

The stored outcome $o_m$ is handled separately. If any supporting evidence
conflicts with the current query, i.e.,
$D_m(q)\cap H_m\neq\emptyset$, \system suppresses the previous outcome.
Otherwise, its use depends on the task. When $\tau_m=\tau_q$, the outcome can
be retained as a previous answer to the same task. When
$\tau_m\neq\tau_q$, \system converts it into an evidence statement labeled by
its original task. For example, a previous answer ``2016'' to a
founding-year query can be represented as
$(\texttt{founded\_in},2016)$ when a later query asks who founded the same
organization. The earlier result can then remain available as contextual
evidence without being presented as the answer to the new question. The
transformed context from one memory is

\begin{equation}
\widetilde{m}(q)
=
U_m(q)
\cup
\begin{cases}
\{o_m\}, &
D_m(q)\cap H_m=\emptyset,\ \tau_m=\tau_q, \\[2pt]
\{\operatorname{Type}(o_m)\}, &
D_m(q)\cap H_m=\emptyset,\ \tau_m\neq\tau_q, \\[2pt]
\emptyset, &
D_m(q)\cap H_m\neq\emptyset ,
\end{cases}
\label{eq:trimem}
\end{equation}

where $\operatorname{Type}(o_m)$ converts the previous outcome into an
evidence unit labeled by its original task. Outcomes that cannot be
meaningfully reused as evidence are omitted.

Finally, \system merges the transformed candidate memories and emits each
retained evidence unit only once. This context is combined with the unchanged
current query and passed to the LLM.


\section{Evaluation}
\label{sec:evaluation}

\begin{table*}[t]
\centering
\scriptsize
\setlength{\tabcolsep}{1.0pt}
\setlength{\aboverulesep}{0pt}
\setlength{\belowrulesep}{0pt}
\renewcommand{\arraystretch}{0.8}
\caption{
Mitigation results (\%). All methods use Mem0 as the memory backbone and
differ only in how retrieved memory is filtered, compressed, or used.
TA denotes task-aligned cases where the stored memory is directly applicable,
and TD-Dup denotes task-divergent cases with duplicated evidence, where the
stored conclusion does not transfer.
\cmark$\rightarrow$\xmark and \xmark$\rightarrow$\cmark count memory-induced
regressions and corrections on TD-Dup cases, respectively.
Standard deviations are shown as subscripts.
}
\begin{tabular*}{\textwidth}{
@{\extracolsep{\fill}}
ll
>{\columncolor{ctrlbg}\color{ctrlfg}}c c
>{\columncolor{flipbad}}c
>{\columncolor{flipgood}}c
>{\columncolor{ctrlbg}\color{ctrlfg}}c c
>{\columncolor{flipbad}}c
>{\columncolor{flipgood}}c
>{\columncolor{ctrlbg}\color{ctrlfg}}c c
>{\columncolor{flipbad}}c
>{\columncolor{flipgood}[\tabcolsep][0pt]}c
@{}
}
\toprule
& &
\multicolumn{4}{c}{\textbf{RippleEdits}} &
\multicolumn{4}{c}{\textbf{CounterLogic}} &
\multicolumn{4}{c}{\textbf{CLADDER-Derived}} \\
\cmidrule(lr){3-6}
\cmidrule(lr){7-10}
\cmidrule(lr){11-14}
\textbf{Model} & \textbf{Method}
& \textbf{TA} & \textbf{TD-Dup} & \ding{51}$\rightarrow$\ding{55} & \ding{55}$\rightarrow$\ding{51}
& \textbf{TA} & \textbf{TD-Dup} & \ding{51}$\rightarrow$\ding{55} & \ding{55}$\rightarrow$\ding{51}
& \textbf{TA} & \textbf{TD-Dup} & \ding{51}$\rightarrow$\ding{55} & \ding{55}$\rightarrow$\ding{51} \\
\midrule
\multirow{12}{*}{Qwen3.8-27B} & No Mem. & \accstd{68.0}{0.4} & \accstd{71.3}{0.8} & -- & -- & \accstd{85.4}{1.3} & \accstd{86.1}{0.9} & -- & -- & \accstd{60.1}{0.5} & \accstd{61.2}{0.9} & -- & -- \\
 & Mem0 & \accstd{85.3}{0.7} & \accstd{62.6}{1.3} & 11.6 & 2.9 & \accstd{93.1}{0.9} & \accstd{80.0}{1.4} & 7.8 & 1.7 & \accstd{79.6}{0.8} & \accstd{46.6}{1.5} & 16.9 & 2.3 \\
\cmidrule(lr){2-14}
 & Facts-only & \accstd{71.6}{0.6} & \accstd{72.4}{0.9} & 2.2 & 3.3 & \accstd{87.0}{1.3} & \accstd{86.7}{0.7} & 1.5 & 2.1 & \accstd{64.1}{1.2} & \accstd{61.0}{1.1} & 3.2 & 3.0 \\
 & Top-1 & \accstd{74.5}{1.2} & \accstd{63.1}{1.5} & 10.8 & 2.6 & \accstd{88.3}{0.5} & \accstd{79.4}{1.2} & 8.3 & 1.6 & \accstd{67.4}{0.6} & \accstd{47.6}{0.7} & 15.7 & 2.1 \\
 & MMR & \accstd{85.7}{1.6} & \accstd{68.9}{1.1} & 5.4 & 3.0 & \accstd{93.3}{0.7} & \accstd{84.3}{0.9} & 3.6 & 1.8 & \accstd{80.2}{1.3} & \accstd{55.7}{1.9} & 7.9 & 2.4 \\
 & Sim. filter & \accstd{75.6}{0.9} & \accstd{66.4}{1.8} & 7.9 & 3.0 & \accstd{88.8}{1.8} & \accstd{82.6}{0.7} & 5.3 & 1.8 & \accstd{68.7}{1.6} & \accstd{52.1}{1.0} & 11.5 & 2.4 \\
 & RECOMP & \accstd{84.1}{1.4} & \accstd{63.1}{1.4} & 11.0 & 2.8 & \accstd{92.6}{0.9} & \accstd{80.4}{1.7} & 7.4 & 1.7 & \accstd{78.2}{0.7} & \accstd{47.4}{1.4} & 16.0 & 2.2 \\
 & LLMLingua-2 & \accstd{79.6}{0.6} & \accstd{60.7}{1.2} & 13.3 & 2.7 & \accstd{90.6}{1.2} & \accstd{78.8}{0.7} & 8.9 & 1.6 & \accstd{73.5}{0.6} & \accstd{44.0}{0.9} & 19.4 & 2.2 \\
 & Context-faithful & \accstd{86.2}{0.5} & \accstd{72.6}{1.4} & 1.4 & 2.7 & \accstd{93.5}{0.9} & \accstd{86.8}{1.4} & 0.9 & 1.6 & \accstd{82.8}{1.1} & \accstd{61.3}{1.0} & 2.0 & 2.1 \\
 & Answer-first & \accstd{77.9}{1.5} & \accstd{71.2}{0.6} & 3.5 & 3.4 & \accstd{89.8}{0.8} & \accstd{85.6}{1.4} & 2.4 & 1.9 & \accstd{71.3}{1.2} & \accstd{58.9}{1.8} & 5.1 & 2.8 \\
 & Micro-Act & \accstd{82.9}{0.6} & \accstd{69.8}{2.3} & 5.0 & 3.5 & \accstd{92.0}{1.8} & \accstd{84.7}{0.8} & 3.4 & 2.0 & \accstd{76.4}{1.0} & \accstd{56.8}{1.7} & 7.3 & 2.9 \\
 & \system~(ours) & \textbf{\accstd{88.4}{1.7}} & \textbf{\accstd{74.5}{1.3}} & \textbf{0.6} & \textbf{3.8} & \textbf{\accstd{94.6}{0.6}} & \textbf{\accstd{87.9}{1.5}} & \textbf{0.4} & \textbf{2.2} & \textbf{\accstd{83.9}{1.5}} & \textbf{\accstd{63.5}{1.4}} & \textbf{0.9} & \textbf{3.2} \\
\midrule
\multirow{12}{*}{GPT-5.6 Luna} & No Mem. & \accstd{65.4}{0.9} & \accstd{68.9}{0.7} & -- & -- & \accstd{84.9}{1.2} & \accstd{84.7}{1.8} & -- & -- & \accstd{58.8}{0.6} & \accstd{58.4}{1.4} & -- & -- \\
 & Mem0 & \accstd{86.1}{0.4} & \accstd{62.2}{1.2} & 9.7 & 3.0 & \accstd{93.4}{0.7} & \accstd{79.4}{1.3} & 7.0 & 1.7 & \accstd{80.4}{0.9} & \accstd{45.5}{1.1} & 15.3 & 2.4 \\
\cmidrule(lr){2-14}
 & Facts-only & \accstd{70.0}{0.9} & \accstd{72.6}{1.6} & 0.9 & 4.6 & \accstd{85.7}{1.3} & \accstd{85.5}{1.2} & 1.3 & 2.1 & \accstd{63.5}{1.7} & \accstd{58.4}{1.3} & 3.1 & 3.1 \\
 & Top-1 & \accstd{73.4}{0.6} & \accstd{62.3}{1.6} & 9.3 & 2.7 & \accstd{88.2}{1.8} & \accstd{79.6}{1.8} & 6.7 & 1.6 & \accstd{66.6}{1.0} & \accstd{46.8}{1.5} & 13.8 & 2.2 \\
 & MMR & \accstd{86.4}{1.1} & \accstd{67.8}{0.8} & 4.2 & 3.1 & \accstd{93.6}{0.6} & \accstd{83.4}{1.7} & 3.1 & 1.8 & \accstd{80.9}{0.8} & \accstd{54.2}{1.1} & 6.7 & 2.5 \\
 & Sim. filter & \accstd{75.0}{0.6} & \accstd{65.0}{1.2} & 7.0 & 3.1 & \accstd{89.7}{1.6} & \accstd{81.7}{1.7} & 4.8 & 1.8 & \accstd{68.8}{0.9} & \accstd{49.9}{1.2} & 11.0 & 2.5 \\
 & RECOMP & \accstd{84.2}{1.6} & \accstd{62.8}{1.9} & 9.0 & 2.9 & \accstd{92.4}{0.7} & \accstd{80.1}{0.9} & 6.3 & 1.7 & \accstd{76.0}{1.1} & \accstd{46.8}{1.4} & 13.9 & 2.3 \\
 & LLMLingua-2 & \accstd{78.8}{0.5} & \accstd{61.0}{1.2} & 10.7 & 2.8 & \accstd{90.5}{1.2} & \accstd{78.4}{1.9} & 7.9 & 1.6 & \accstd{73.7}{1.2} & \accstd{42.6}{1.5} & 18.1 & 2.3 \\
 & Context-faithful & \accstd{87.0}{0.6} & \accstd{70.5}{1.9} & 1.2 & 2.8 & \accstd{93.8}{1.6} & \accstd{85.4}{1.7} & 0.9 & 1.6 & \accstd{81.6}{1.0} & \accstd{58.8}{0.7} & 1.8 & 2.2 \\
 & Answer-first & \accstd{77.5}{0.6} & \accstd{69.4}{0.7} & 3.0 & 3.5 & \accstd{86.5}{0.7} & \accstd{84.6}{1.1} & 2.0 & 1.9 & \accstd{70.3}{0.5} & \accstd{61.8}{0.8} & 1.4 & 4.8 \\
 & Micro-Act & \accstd{82.1}{1.0} & \accstd{68.6}{0.6} & 3.9 & 3.6 & \accstd{91.7}{1.3} & \accstd{83.5}{0.8} & 3.2 & 2.0 & \accstd{79.7}{1.0} & \accstd{55.1}{1.1} & 6.3 & 3.0 \\
 & \system~(ours) & \textbf{\accstd{89.0}{0.7}} & \textbf{\accstd{75.3}{1.8}} & \textbf{0.5} & \textbf{6.9} & \textbf{\accstd{94.8}{1.8}} & \textbf{\accstd{88.5}{1.3}} & \textbf{0.4} & \textbf{4.2} & \textbf{\accstd{84.5}{1.1}} & \textbf{\accstd{64.9}{0.7}} & \textbf{0.8} & \textbf{7.3} \\
\bottomrule
\end{tabular*}
\label{tab:mitigation}
\end{table*}

We evaluate whether \system reduces memory overreliance while preserving the
benefit of useful memory. We detail settings and evaluations as below:

\textbf{Evaluation settings.}
To make \system directly
comparable with the previous results. We use the same benchmarks, backbone LLMs, memory systems, and metrics as in
\S\ref{sec:overreliance}. 

\textbf{Baselines:}
We compare against several representative strategies applied on top of each
agentic memory system: (i) \emph{Facts-only} removes the stored answer and
retains only factual content; (ii) \emph{Top-1} keeps only the highest-ranked
retrieved memory; (iii) MMR~\citep{carbonell1998use} reranks memories to reduce
redundancy; (iv) \emph{Similarity Filter} removes memories below a similarity
threshold; (v) RECOMP~\citep{xu2024recomp} compresses retrieved memories into
task-relevant evidence; (vi) LLMLingua-2~\citep{pan2024llmlingua} shortens the
retrieved context through prompt compression; (vii) \emph{Context-faithful}
~\citep{zhou2023context} prompts the model to prioritize the current context
under retrieval conflict; (viii) \emph{Answer-first}
~\citep{neeman2023disentqa} answers from the current query before comparing
against retrieved information; and (ix) Micro-Act~\citep{huo2025micro} adds an
explicit reasoning step for handling conflicting retrieved knowledge.

\subsection{Does \system mitigate overreliance while preserving useful memory?}

We first compare all methods using the same underlying retrieved memories. The
main question is whether a method can improve reasoning on TD-Dup cases while
preserving the benefits of memory on TA cases. Table~\ref{tab:mitigation} shows that \system improves this trade-off. On
TD-Dup cases, it recovers much of the performance lost under unmodified
memory, while TA performance remains close to (or even better than) the original memory system. The
same trend appears for different LLM backbones and extends to the other memory
architectures in the Appendix \ref{app:other-memory}. Hence, we summarize that \system reduces incorrect memory-based inference while retaining (or improving) cases where memory is helpful given the structured representation in \system.

The comparison also clarifies why simply weakening memory is insufficient.
Baselines that remove entries, discard stored answers, or compress the retrieved
context can reduce harmful influence on TD-Dup cases, but can also discard
information that remains useful in TA cases. Prompting-based methods preserve
more of the original context, yet overlapping evidence can still reinforce a
conclusion that does not transfer to the current task. \system
works at the evidence level, controlling replicated or conflicting memory evidence
while retaining memory-specific information that remains useful.

\subsection{Which parts of \system are responsible for the mitigation?}

\begin{wrapfigure}{r}{0.42\textwidth}
    \vspace{-10pt}
    \centering
    \includegraphics[width=0.42\textwidth]{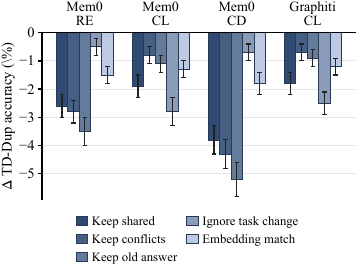}
    \vspace{-10pt}
\caption{
Ablation of \system on TD-Dup cases with Qwen3.8-27B.
Bars show the accuracy change relative to the full method; lower values
indicate a larger drop. RE, CL, and CD denote RippleEdits, CounterLogic,
and CLADDER-Derived.
}
    \label{fig:ablation}
    \vspace{-10pt}
\end{wrapfigure}

We ablate the main read-time steps in \system.
\emph{Keep shared} retains evidence already present in the current query;
\emph{Keep conflicts} retains memory evidence that conflicts with the query;
\emph{Keep old answer} keeps the previous answer even when the evidence it
relied on has changed; \emph{Ignore task change} directly reuses the previous
answer when the current query asks a different question; and
\emph{Embedding match} replaces trie-based evidence matching with embedding
similarity.

Figure~\ref{fig:ablation} shows that these steps matter in different settings.
Keeping shared evidence consistently lowers TD-Dup accuracy, showing that
repeated evidence contributes to overreliance. Keeping conflicting evidence
or the old answer has a larger effect on RippleEdits and CLADDER-Derived,
where answer-relevant information changes. Ignoring the task change matters
most on CounterLogic, where much of the evidence stays the same but the
question changes. Embedding matching causes a smaller but consistent drop.

Overall, the ablations follow the design of \system: removing repeated
evidence reduces duplicated influence, removing conflicts keeps memory
consistent with the current query, dropping an outdated answer prevents it
from carrying over after its evidence changes, and distinguishing task changes
prevents an answer to one question from being reused for another.

\subsection{Does \system remain effective when memory is attacked?}

\begin{figure*}[t]
    \centering
    \includegraphics[width=\textwidth]{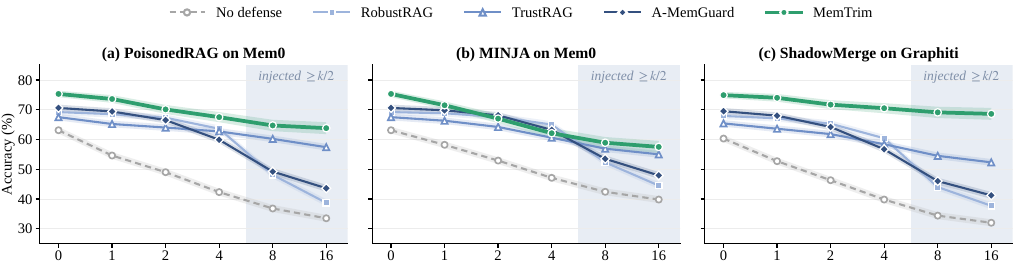}
    \caption{
    TD-Dup accuracy as misleading memory is increasingly injected into the
    retrieved context. The shaded region marks settings where injected entries
    form a large fraction of retrieval. Bands show confidence intervals across
    runs.
    }
    \label{fig:attack}
\end{figure*}

Our main experiments use memories formed through ordinary interaction. We
further test \system under deliberate memory attacks. PoisonedRAG~\citep{zou2025poisonedrag}
injects repeated stale records, MINJA~\citep{dong2026memory} induces the agent
to store such records through normal queries, and
ShadowMerge~\citep{luo2026shadowmerge} inserts conflicting relations into
graph memory. We increase the number of injected entries while keeping the
retrieval budget fixed, and compare \system with
RobustRAG~\citep{xiang2024certifiably}, TrustRAG~\citep{zhou2025trustrag}, and
A-MemGuard~\citep{wei2025memguard}.

Figure~\ref{fig:attack} shows that all methods degrade as attacked memories
occupy more of the retrieved context, but \system degrades more slowly. The
difference becomes clearer as injected entries make up a larger share of the
retrieved memories, and the same pattern appears across both text-based and
graph-based attacks. The result highlights a useful property of \system: repeated stale memories do
not translate directly into repeated influence on the model. \system removes
duplicated evidence and excludes conflicting information before inference, so
injecting more copies has a weaker effect on the final context. This suggests
that the same design used to reduce benign memory overreliance also provides
some resistance when misleading memory is deliberately amplified.

\begin{figure*}[t]
    \centering
    \includegraphics[width=\textwidth]{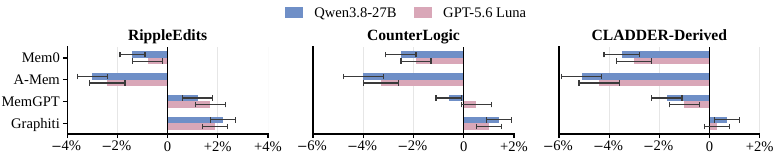}
    \caption{
    Change in total reasoning time after adding \system to each memory system.
    Negative values indicate shorter reasoning time. Error bars show variation
    across runs.
    }
    \label{fig:overhead}
\end{figure*}

\subsection{Does \system Introduce Significant Running Overhead?}

We measure total reasoning time after adding \system, including memory reading
and answer generation; stored memories are processed once at write time and
reused across queries. Figure~\ref{fig:overhead} shows only small runtime
changes across memory systems. \system adds some cost when matching evidence,
but it can also shorten the retrieved context before generation, which offsets
part of this cost and can even reduce runtime in some settings. Overall,
\system improves memory use without substantially increasing inference time.




\vspace{-5pt}
\section{Related Work}
\label{sec:related}

\textbf{Memory in LLM agents.}
Memory systems let an agent keep and reuse past experience. One line of works build
the storage: long-term memory across
sessions~\citep{park2023generative,zhong2024memorybank}, context
hierarchies~\citep{packer2023memgpt}, associative
indexes~\citep{gutierrez2024hipporag}, and self-organized
memory~\citep{xu2026mem}. Other works turn past trajectories into reusable
knowledge such as reflections, lessons, and
workflows~\citep{shinn2023reflexion,zhao2024expel,wang2024agent}, and recent
benchmarks evaluate these systems in multi-turn
interaction~\citep{hu2026evaluating}. All of these designs leverage
retrieved relevant memory to help the current reasoning tasks.

\textbf{Failures and attacks on memory reuse.}
Memory can also hurt. Agents copy retrieved experiences with similar inputs,
so past errors spread~\citep{xiong-etal-2026-memory}. Memories are applied
where they no longer fit~\citep{pulipaka2026persistbench,piehl2026er}, turn faulty under
repeated updates~\citep{zhang2026useful,meng2026equimem}, or look similar to the query while
being wrong for the task~\citep{tang2026trap,zhang2026beyond}. Memory is also an attack
surface: adversaries poison memory banks~\citep{chen2024agentpoison},
retrieval corpora~\citep{zou2025poisonedrag}, or self-evolving
experiences~\citep{wang2026oep}, and defenses curate data against such
samples~\citep{liu2024data}. In all these cases, the failure
comes from a flawed memory, a flawed retrieval, or an attacker. However, memory
overreliance needs none of them: a correct, well-retrieved memory fails
simply because its answer-relevant condition has changed.

\textbf{Reasoning under varying conditions.} Knowledge editing
asks how models handle updated facts and their ripple
effects~\citep{meng2022locating,cohen2024evaluating}. Counterfactual
reasoning asks how models handle changed premises, with benchmarks on causal
interventions~\citep{yang2026eligibility,jin2023cladder}, alternative story
questions~\citep{ashida2022possible}, and knowledge-conflicting
logic~\citep{joishy2025flying}, and with analyses of where LLMs
fail~\citep{tang2026value,youssef2024llms,yang2026eligibility}. Such changes are routine in
practice: in cyber threat intelligence, the landscape shifts fast enough that
stored assessments go stale~\citep{meng2025uncovering,meng2025benchmarking,tang2025polar}.
Prior work treats a changed condition as a reasoning challenge for the model.
We show that memory makes it worse: the stored answer anchors the model to
the old condition, and our framework removes this anchor at read time.


\vspace{-5pt}
\section{Conclusion}
\vspace{-3pt}

We study memory overreliance, where benign and appropriately retrieved memory
can misguide inference when only part of the stored experience transfers to
the current task. Our experiments identify partial evidence overlap as the
most vulnerable regime and motivate \system, which controls repeated and
conflicting memory evidence before inference while preserving useful
memory-specific information. Across models and memory architectures, \system
reduces harmful memory reuse while retaining the benefits of applicable
memory, with small runtime changes and added resistance under memory attacks.

\bibliographystyle{iclr2027_conference}
\bibliography{iclr2027_conference}


\appendix

\section{Complementary Technical Details of \system}

\subsection{Implementation Details}
\label{app:implementation}

This part provides implementation details omitted from
\S\ref{sec:method}. We focus on how free-text interactions are converted into
stable evidence units, how answer-support sets and task signatures are
constructed, and how evidence from multiple retrieved memories is combined at
read time.

\textbf{Canonicalizing evidence units.}
The same fact can be expressed in many ways, so \system does not index raw
text directly. For each interaction, we use a fixed parser prompt to extract
typed evidence records. We retain the notation $e=(k,v)$ from
\S\ref{ssec:method-overview}, but internally the key $k$ contains the entity
being described, its attribute or relation, and its value type. For example,
``Highway~5 is closed,'' ``Route~5 has been shut,'' and ``Highway~5 is not
open'' can all map to
$(\texttt{Highway5\_status},\texttt{closed})$.
In contrast, the word ``open'' in ``the restaurant is open'' is assigned a
different key because it refers to a different entity and attribute. This
prevents surface similarity alone from merging evidence with different
meanings.

When a new unit is parsed, \system first checks for an exact canonical key in
the trie. If none exists, it retrieves a small set of existing keys with the
same entity and value type and asks the parser whether the new key is
semantically equivalent to any of them. Equivalent keys are merged; otherwise
a new key is created. Values are aligned only after their keys match. Boolean,
categorical, numeric, and temporal values are normalized into a fixed form;
free-text values use the same equivalence check. When equivalence is unclear,
\system keeps the units separate. This conservative rule avoids merging two
pieces of evidence only because they share similar wording.

Each parsed unit also keeps a pointer to the text span from which it was
extracted. We use this span to verify that every indexed unit is grounded in
the original interaction. A parsed unit that cannot be traced to the input is
discarded before insertion into the trie.

\begin{casebox}{\bf Canonical Evidence Alignment}
Suppose one memory states \emph{``Highway~5 is closed today''}, while another
states \emph{``Route~5 has been shut for the day.''} Both are normalized to
$(\texttt{Highway5\_status},\texttt{closed})$ and therefore share one trie
node. If a third memory states \emph{``Highway~5 is open''}, it shares the same
key but follows a different value branch,
$\texttt{Highway5\_status=open}$.

The word \emph{open} may also occur in an unrelated sentence such as
\emph{``the visitor center is open.''} This evidence is indexed under a
different key, e.g.,
$(\texttt{VisitorCenter\_status},\texttt{open})$, and is never merged with
the highway-status evidence.
\end{casebox}

\textbf{Extracting task signatures and answer support.}
The parser also produces the task signature $\tau_m$ and support set $H_m$.
A task signature records both the requested operation and its target. For
example, ``Which route should I take?'' and ``Recommend a route to City~B''
map to the same \texttt{route\_planning} signature, whereas ``How long will
the trip take?'' maps to \texttt{travel\_time}. We canonicalize task
signatures with the same procedure used for evidence keys. If two task
descriptions cannot be confidently aligned, we treat them as different tasks;
this prevents an earlier answer from being reused under an uncertain task
match.

To construct $H_m$, the parser marks which evidence units directly support
the stored outcome $o_m$. For the route-planning example, the answer
``Take Highway~5'' may depend on
$(\texttt{Highway5\_status},\texttt{open})$ and
$(\texttt{destination},B)$, while an unrelated fact such as the weather need
not belong to $H_m$. We use a conservative policy for uncertain cases:
evidence is included in $H_m$ when the parser cannot determine whether the
stored outcome depends on it. This can suppress reuse more often, but avoids
keeping an outcome after potentially important support has changed.

When $\tau_m\neq\tau_q$, \system converts a previous outcome into evidence
only when it can be expressed as a factual key-value unit without changing
its meaning. For example, ``Take Highway~5'' can be converted to
$(\texttt{recommended\_route},\texttt{Highway5})$ when a later query asks
about travel time. Outputs that are inherently procedural, subjective, or
cannot be represented faithfully as evidence are omitted under a task change.

\textbf{Combining multiple retrieved memories.}
The trie is used only to process memories already returned by the underlying
retriever. Given $\mathcal{R}(q)$, \system restricts all trie matches to the
memory identifiers in this set; it does not use the trie as an additional
retrieval system.

After each retrieved memory is transformed as described in
\S\ref{ssec:method-read}, \system merges the retained evidence using its
canonical trie identifiers. Identical memory-only evidence from several
memories is emitted once. If several retrieved memories provide different
values for the same key and the current query does not resolve the conflict,
\system does not select a value based only on retrieval similarity. By
default, all unresolved values for that key are withheld from the final
context. If the base memory system provides an explicit ordering signal such
as a trusted timestamp, that metadata can be used to resolve the conflict
before this step.

The current query always takes precedence when it provides a value for the
same key. Thus, if the stored memory contains
$\texttt{Highway5\_status=open}$ while the current query contains
$\texttt{Highway5\_status=closed}$, the stored value is removed regardless of
the memory's retrieval score.

\textbf{Parsing failures and runtime cost.}
Memory evidence is parsed once when the memory is written, and its structured
representation is cached with the trie index. At inference time, \system only
parses the current query once; retrieved memories reuse their cached evidence
units, task signatures, and support sets. Trie matching therefore consists of
key-value lookups over the evidence units in $E_q$ and the retrieved memory
identifiers.

We use constrained structured output and deterministic decoding for the
parser. Each result is checked for valid fields, grounded source spans, and
well-formed key-value types. If parsing fails, we retry once with the same
interaction and the validation error. If a valid representation still cannot
be produced, the raw memory remains in the base memory store but its stored
outcome is not reused through \system. This keeps parser failures from
introducing an unchecked answer into the inference context.

\section{Additional Results}
\label{app:other-dataset}

This appendix extends the main results in three directions.
Appendices~\ref{app:other-dataset-rq2} and~\ref{app:other-dataset-rq3} repeat
RQ2 and RQ3 of Section~\ref{sec:overreliance} on CounterLogic.
Appendix~\ref{app:other-memory} repeats the mitigation comparison of
Section~\ref{sec:evaluation} on the other three memory systems.
Appendix~\ref{app:trimem-cases} walks through one case from each dataset to
show how partial evidence overlap misleads ordinary memory use and how
\system corrects it.

\subsection{RQ2 on CounterLogic: Where Failures Occur}
\label{app:other-dataset-rq2}

\begin{figure}[h]
\centering
\includegraphics[width=\linewidth]{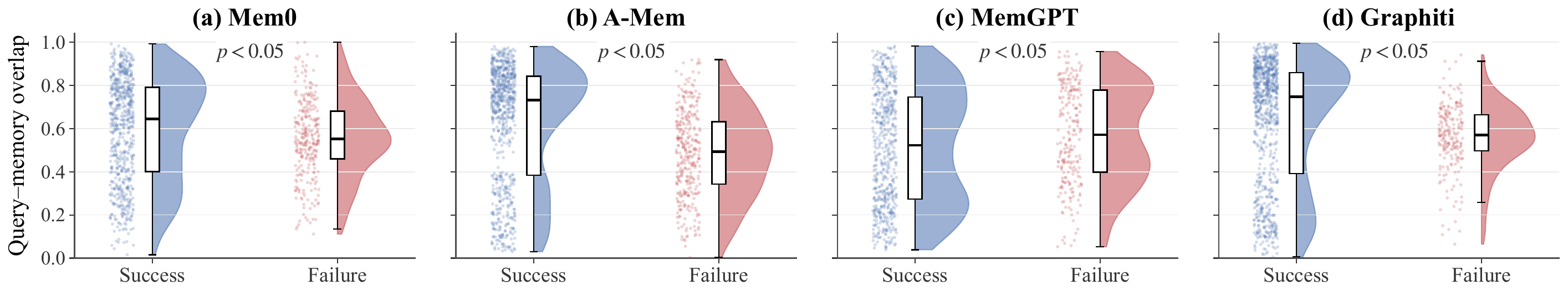}
\captionsetup{font=footnotesize}
\caption{
Query overlap in successful and failed cases on CounterLogic, for each memory
system.
}
\label{fig:rq2-counterlogic}
\end{figure}

Figure~\ref{fig:rq2-counterlogic} shows the same separation as
Figure~\ref{fig:rq2}. Successful cases occur more often near the two ends of
the overlap range, while failed cases concentrate at partial overlap.

\subsection{RQ3 on CounterLogic: Controlled Duplication}
\label{app:other-dataset-rq3}

\begin{figure}[h]
\centering
\includegraphics[width=\linewidth]{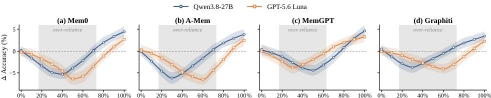}
\captionsetup{font=footnotesize}
\caption{
Accuracy change across levels of duplicated evidence on CounterLogic, for each
memory system.
}
\label{fig:rq3-counterlogic}
\end{figure}

Figure~\ref{fig:rq3-counterlogic} shows the same non-monotonic pattern as
Figure~\ref{fig:rq3}. The accuracy loss is largest at intermediate duplication
and disappears as duplication approaches full overlap. The drop is shallower
than on RippleEdits, and the gain at full duplication is smaller.

\subsection{Mitigation on Other Memory Systems}
\label{app:other-memory}

Table~\ref{tab:mitigation} fixes Mem0 as the memory backbone so that its rows
differ only in the defense. Tables~\ref{tab:other-memory-qwen}
and~\ref{tab:other-memory-gpt} replace Mem0 with A-Mem, MemGPT, and
Graphiti, and keep the same baselines, datasets, and columns. The no-memory
baseline does not depend on the memory system and is reported in
Table~\ref{tab:rq1}.

\begin{table*}[t]
\centering
\scriptsize
\setlength{\tabcolsep}{1.0pt}
\setlength{\aboverulesep}{0pt}
\setlength{\belowrulesep}{0pt}
\renewcommand{\arraystretch}{0.8}
\caption{
Mitigation results (\%) with Qwen3.8-27B on the other three memory systems.
Each block uses the named system as the memory backbone, and its first two rows
match Table~\ref{tab:rq1}.
TA and TD-Dup follow Table~\ref{tab:mitigation}, and
\cmark$\rightarrow$\xmark and \xmark$\rightarrow$\cmark are counted on TD-Dup cases.
}
\begin{tabular*}{\textwidth}{
@{\extracolsep{\fill}}
ll
>{\columncolor{ctrlbg}\color{ctrlfg}}c c
>{\columncolor{flipbad}}c
>{\columncolor{flipgood}}c
>{\columncolor{ctrlbg}\color{ctrlfg}}c c
>{\columncolor{flipbad}}c
>{\columncolor{flipgood}}c
>{\columncolor{ctrlbg}\color{ctrlfg}}c c
>{\columncolor{flipbad}}c
>{\columncolor{flipgood}}c
@{}
}
\toprule
& &
\multicolumn{4}{c}{\textbf{RippleEdits}} &
\multicolumn{4}{c}{\textbf{CounterLogic}} &
\multicolumn{4}{c}{\textbf{CLADDER-Derived}} \\
\cmidrule(lr){3-6}
\cmidrule(lr){7-10}
\cmidrule(lr){11-14}
\textbf{Memory} & \textbf{Method}
& \textbf{TA} & \textbf{TD-Dup} & \cmark$\rightarrow$\xmark & \xmark$\rightarrow$\cmark
& \textbf{TA} & \textbf{TD-Dup} & \cmark$\rightarrow$\xmark & \xmark$\rightarrow$\cmark
& \textbf{TA} & \textbf{TD-Dup} & \cmark$\rightarrow$\xmark & \xmark$\rightarrow$\cmark \\
\midrule
\multirow{12}{*}{A-Mem} & No Mem. & \accstd{68.0}{0.4} & \accstd{71.3}{0.8} & -- & -- & \accstd{85.4}{1.3} & \accstd{86.1}{0.9} & -- & -- & \accstd{60.1}{0.5} & \accstd{61.2}{0.9} & -- & -- \\
 & A-Mem & \accstd{82.6}{1.0} & \accstd{61.2}{1.5} & 12.8 & 2.7 & \accstd{92.4}{0.6} & \accstd{80.6}{1.2} & 7.1 & 1.6 & \accstd{78.4}{0.9} & \accstd{46.1}{1.4} & 17.6 & 2.5 \\
\cmidrule(lr){2-14}
 & \quad Facts-only & \accstd{71.0}{0.5} & \accstd{72.0}{0.4} & 2.4 & 3.1 & \accstd{86.8}{0.8} & \accstd{86.8}{1.4} & 1.3 & 2.0 & \accstd{63.6}{1.1} & \accstd{61.3}{1.9} & 3.1 & 3.2 \\
 & \quad Top-1 & \accstd{73.5}{2.0} & \accstd{61.7}{0.6} & 12.0 & 2.4 & \accstd{88.2}{1.8} & \accstd{80.5}{1.7} & 7.1 & 1.5 & \accstd{66.7}{1.6} & \accstd{47.7}{0.8} & 15.8 & 2.3 \\
 & \quad MMR & \accstd{83.1}{1.5} & \accstd{67.9}{1.3} & 6.2 & 2.8 & \accstd{92.6}{0.7} & \accstd{84.6}{2.1} & 3.2 & 1.7 & \accstd{79.0}{0.6} & \accstd{55.0}{1.0} & 8.8 & 2.6 \\
 & \quad Sim. filter & \accstd{74.7}{0.9} & \accstd{64.9}{1.0} & 9.2 & 2.8 & \accstd{88.5}{1.2} & \accstd{83.0}{2.2} & 4.8 & 1.7 & \accstd{67.5}{1.8} & \accstd{52.7}{1.9} & 11.1 & 2.6 \\
 & \quad RECOMP & \accstd{82.0}{0.7} & \accstd{61.4}{1.1} & 12.5 & 2.6 & \accstd{92.4}{1.3} & \accstd{80.5}{1.9} & 7.2 & 1.6 & \accstd{76.8}{1.7} & \accstd{47.3}{1.4} & 16.3 & 2.4 \\
 & \quad LLMLingua-2 & \accstd{77.3}{1.2} & \accstd{59.8}{0.7} & 14.0 & 2.5 & \accstd{90.4}{1.8} & \accstd{78.9}{0.6} & 8.7 & 1.5 & \accstd{72.7}{1.1} & \accstd{42.9}{0.5} & 20.7 & 2.4 \\
 & \quad Context-faithful & \accstd{83.6}{2.0} & \accstd{72.3}{2.3} & 1.5 & 2.5 & \accstd{92.8}{1.3} & \accstd{86.8}{2.1} & 0.8 & 1.5 & \accstd{79.6}{0.9} & \accstd{61.5}{1.2} & 2.0 & 2.3 \\
 & \quad Answer-first & \accstd{77.0}{0.5} & \accstd{70.3}{0.6} & 4.2 & 3.2 & \accstd{89.7}{2.2} & \accstd{85.6}{1.8} & 2.3 & 1.8 & \accstd{69.9}{1.5} & \accstd{59.2}{0.8} & 5.0 & 3.0 \\
 & \quad Micro-Act & \accstd{81.2}{2.3} & \accstd{68.8}{1.6} & 5.8 & 3.3 & \accstd{91.9}{1.3} & \accstd{84.7}{1.2} & 3.3 & 1.9 & \accstd{76.0}{1.8} & \accstd{56.7}{0.7} & 7.6 & 3.1 \\
 & \quad \system~(ours) & \textbf{\accstd{86.3}{0.5}} & \textbf{\accstd{74.2}{0.7}} & \textbf{0.7} & \textbf{3.6} & \textbf{\accstd{94.0}{1.6}} & \textbf{\accstd{87.8}{2.3}} & \textbf{0.4} & \textbf{2.1} & \textbf{\accstd{83.0}{1.9}} & \textbf{\accstd{63.7}{1.0}} & \textbf{0.9} & \textbf{3.4} \\
\midrule
\multirow{12}{*}{MemGPT} & No Mem. & \accstd{68.0}{0.4} & \accstd{71.3}{0.8} & -- & -- & \accstd{85.4}{1.3} & \accstd{86.1}{0.9} & -- & -- & \accstd{60.1}{0.5} & \accstd{61.2}{0.9} & -- & -- \\
 & MemGPT & \accstd{80.9}{0.8} & \accstd{65.7}{1.1} & 8.9 & 3.3 & \accstd{91.2}{1.1} & \accstd{82.8}{1.0} & 5.2 & 1.9 & \accstd{78.1}{0.7} & \accstd{49.8}{1.2} & 14.1 & 2.7 \\
\cmidrule(lr){2-14}
 & \quad Facts-only & \accstd{70.5}{2.2} & \accstd{73.4}{1.4} & 1.6 & 3.7 & \accstd{86.6}{2.1} & \accstd{87.4}{1.2} & 1.0 & 2.3 & \accstd{63.8}{1.0} & \accstd{62.0}{0.7} & 2.6 & 3.4 \\
 & \quad Top-1 & \accstd{73.2}{1.2} & \accstd{65.5}{1.5} & 8.8 & 3.0 & \accstd{87.6}{2.0} & \accstd{83.0}{2.3} & 4.9 & 1.8 & \accstd{67.3}{1.3} & \accstd{49.7}{0.8} & 14.0 & 2.5 \\
 & \quad MMR & \accstd{81.4}{1.3} & \accstd{70.5}{1.6} & 4.2 & 3.4 & \accstd{91.4}{0.4} & \accstd{85.8}{0.9} & 2.3 & 2.0 & \accstd{78.7}{1.4} & \accstd{57.8}{2.2} & 6.2 & 2.8 \\
 & \quad Sim. filter & \accstd{73.7}{1.1} & \accstd{68.6}{1.0} & 6.1 & 3.4 & \accstd{88.0}{1.5} & \accstd{84.5}{0.6} & 3.6 & 2.0 & \accstd{68.1}{1.9} & \accstd{54.3}{1.6} & 9.7 & 2.8 \\
 & \quad RECOMP & \accstd{80.5}{1.0} & \accstd{65.7}{0.4} & 8.8 & 3.2 & \accstd{91.0}{0.9} & \accstd{82.9}{1.1} & 5.1 & 1.9 & \accstd{77.2}{1.6} & \accstd{50.2}{1.4} & 13.6 & 2.6 \\
 & \quad LLMLingua-2 & \accstd{76.9}{1.3} & \accstd{63.9}{2.2} & 10.5 & 3.1 & \accstd{89.3}{1.2} & \accstd{82.0}{2.1} & 5.9 & 1.8 & \accstd{72.9}{1.1} & \accstd{46.7}{1.5} & 17.1 & 2.6 \\
 & \quad Context-faithful & \accstd{82.0}{1.8} & \accstd{73.4}{2.0} & 1.0 & 3.1 & \accstd{91.7}{1.3} & \accstd{87.3}{0.6} & 0.6 & 1.8 & \accstd{79.4}{1.9} & \accstd{62.0}{0.4} & 1.7 & 2.5 \\
 & \quad Answer-first & \accstd{75.6}{2.0} & \accstd{72.3}{2.1} & 2.8 & 3.8 & \accstd{88.5}{1.6} & \accstd{86.7}{0.7} & 1.5 & 2.1 & \accstd{69.8}{1.2} & \accstd{60.4}{1.9} & 4.0 & 3.2 \\
 & \quad Micro-Act & \accstd{78.6}{1.9} & \accstd{71.5}{0.4} & 3.7 & 3.9 & \accstd{90.4}{0.5} & \accstd{86.1}{1.1} & 2.2 & 2.2 & \accstd{74.8}{0.9} & \accstd{58.8}{0.7} & 5.7 & 3.3 \\
 & \quad \system~(ours) & \textbf{\accstd{84.9}{0.7}} & \textbf{\accstd{75.0}{1.8}} & \textbf{0.5} & \textbf{4.2} & \textbf{\accstd{93.1}{1.6}} & \textbf{\accstd{88.2}{0.6}} & \textbf{0.3} & \textbf{2.4} & \textbf{\accstd{82.7}{0.5}} & \textbf{\accstd{64.1}{1.5}} & \textbf{0.7} & \textbf{3.6} \\
\midrule
\multirow{12}{*}{Graphiti} & No Mem. & \accstd{68.0}{0.4} & \accstd{71.3}{0.8} & -- & -- & \accstd{85.4}{1.3} & \accstd{86.1}{0.9} & -- & -- & \accstd{60.1}{0.5} & \accstd{61.2}{0.9} & -- & -- \\
 & Graphiti & \accstd{87.4}{0.5} & \accstd{58.3}{1.7} & 15.4 & 2.4 & \accstd{94.6}{0.7} & \accstd{78.7}{1.6} & 8.9 & 1.5 & \accstd{81.2}{0.6} & \accstd{43.9}{1.8} & 19.4 & 2.1 \\
\cmidrule(lr){2-14}
 & \quad Facts-only & \accstd{71.9}{2.3} & \accstd{71.3}{1.8} & 2.8 & 2.8 & \accstd{87.2}{1.1} & \accstd{86.4}{1.5} & 1.6 & 1.9 & \accstd{64.1}{1.4} & \accstd{60.6}{0.6} & 3.4 & 2.8 \\
 & \quad Top-1 & \accstd{74.8}{1.1} & \accstd{60.0}{1.9} & 13.4 & 2.1 & \accstd{89.0}{1.5} & \accstd{78.8}{2.2} & 8.7 & 1.4 & \accstd{67.9}{2.1} & \accstd{45.3}{2.0} & 17.8 & 1.9 \\
 & \quad MMR & \accstd{87.7}{1.6} & \accstd{66.9}{1.7} & 6.9 & 2.5 & \accstd{94.8}{2.3} & \accstd{83.5}{0.6} & 4.2 & 1.6 & \accstd{81.7}{1.5} & \accstd{55.0}{0.4} & 8.4 & 2.2 \\
 & \quad Sim. filter & \accstd{75.9}{2.1} & \accstd{64.0}{1.5} & 9.8 & 2.5 & \accstd{89.6}{0.5} & \accstd{81.3}{1.8} & 6.4 & 1.6 & \accstd{68.7}{1.4} & \accstd{51.1}{1.9} & 12.3 & 2.2 \\
 & \quad RECOMP & \accstd{85.6}{1.8} & \accstd{59.4}{1.4} & 14.2 & 2.3 & \accstd{93.6}{1.6} & \accstd{79.5}{1.3} & 8.1 & 1.5 & \accstd{78.9}{1.9} & \accstd{45.6}{0.8} & 17.6 & 2.0 \\
 & \quad LLMLingua-2 & \accstd{80.1}{0.4} & \accstd{57.1}{0.5} & 16.4 & 2.2 & \accstd{91.4}{1.0} & \accstd{77.6}{1.8} & 9.9 & 1.4 & \accstd{74.2}{0.8} & \accstd{41.0}{1.2} & 22.2 & 2.0 \\
 & \quad Context-faithful & \accstd{88.2}{0.7} & \accstd{71.6}{1.5} & 1.9 & 2.2 & \accstd{94.9}{2.2} & \accstd{86.5}{0.5} & 1.0 & 1.4 & \accstd{82.3}{1.0} & \accstd{60.8}{1.6} & 2.3 & 1.9 \\
 & \quad Answer-first & \accstd{79.7}{2.0} & \accstd{69.3}{1.6} & 4.9 & 2.9 & \accstd{90.9}{0.6} & \accstd{85.0}{2.2} & 2.8 & 1.7 & \accstd{72.5}{1.9} & \accstd{57.8}{2.1} & 6.0 & 2.6 \\
 & \quad Micro-Act & \accstd{85.6}{1.9} & \accstd{67.3}{1.3} & 7.0 & 3.0 & \accstd{93.2}{1.2} & \accstd{84.1}{1.8} & 3.8 & 1.8 & \accstd{77.2}{0.9} & \accstd{56.0}{0.6} & 7.9 & 2.7 \\
 & \quad \system~(ours) & \textbf{\accstd{90.1}{0.7}} & \textbf{\accstd{73.8}{2.2}} & \textbf{0.8} & \textbf{3.3} & \textbf{\accstd{95.7}{1.4}} & \textbf{\accstd{87.6}{1.6}} & \textbf{0.5} & \textbf{2.0} & \textbf{\accstd{85.2}{1.5}} & \textbf{\accstd{63.2}{1.3}} & \textbf{1.0} & \textbf{3.0} \\
\bottomrule
\end{tabular*}
\label{tab:other-memory-qwen}
\end{table*}

\begin{table*}[t]
\centering
\scriptsize
\setlength{\tabcolsep}{1.5pt}
\setlength{\aboverulesep}{0.5pt}
\setlength{\belowrulesep}{0.5pt}
\renewcommand{\arraystretch}{1.0}
\caption{
Mitigation results (\%) with GPT-5.6 Luna on the other three memory systems.
Each block uses the named system as the memory backbone, and its first two rows
match Table~\ref{tab:rq1}.
TA and TD-Dup follow Table~\ref{tab:mitigation}, and
\cmark$\rightarrow$\xmark and \xmark$\rightarrow$\cmark are counted on TD-Dup cases.
}
\begin{tabular*}{\textwidth}{
@{\extracolsep{\fill}}
ll
>{\columncolor{ctrlbg}\color{ctrlfg}}c c
>{\columncolor{flipbad}}c
>{\columncolor{flipgood}}c
>{\columncolor{ctrlbg}\color{ctrlfg}}c c
>{\columncolor{flipbad}}c
>{\columncolor{flipgood}}c
>{\columncolor{ctrlbg}\color{ctrlfg}}c c
>{\columncolor{flipbad}}c
>{\columncolor{flipgood}}c
@{}
}
\toprule
& &
\multicolumn{4}{c}{\textbf{RippleEdits}} &
\multicolumn{4}{c}{\textbf{CounterLogic}} &
\multicolumn{4}{c}{\textbf{CLADDER-Derived}} \\
\cmidrule(lr){3-6}
\cmidrule(lr){7-10}
\cmidrule(lr){11-14}
\textbf{Memory} & \textbf{Method}
& \textbf{TA} & \textbf{TD-Dup} & \cmark$\rightarrow$\xmark & \xmark$\rightarrow$\cmark
& \textbf{TA} & \textbf{TD-Dup} & \cmark$\rightarrow$\xmark & \xmark$\rightarrow$\cmark
& \textbf{TA} & \textbf{TD-Dup} & \cmark$\rightarrow$\xmark & \xmark$\rightarrow$\cmark \\
\midrule
\multirow{12}{*}{A-Mem} & No Mem. & \accstd{65.4}{0.9} & \accstd{68.9}{0.7} & -- & -- & \accstd{84.9}{1.2} & \accstd{84.7}{1.8} & -- & -- & \accstd{58.8}{0.6} & \accstd{58.4}{1.4} & -- & -- \\
 & A-Mem & \accstd{83.7}{1.0} & \accstd{60.8}{1.1} & 10.9 & 2.8 & \accstd{92.2}{0.9} & \accstd{80.1}{1.0} & 6.4 & 1.8 & \accstd{78.9}{1.1} & \accstd{46.2}{1.6} & 14.8 & 2.6 \\
\cmidrule(lr){2-14}
 & \quad Facts-only & \accstd{69.1}{0.5} & \accstd{70.1}{1.5} & 2.0 & 3.2 & \accstd{86.4}{2.2} & \accstd{85.7}{1.1} & 1.2 & 2.2 & \accstd{62.9}{1.7} & \accstd{59.0}{0.4} & 2.7 & 3.3 \\
 & \quad Top-1 & \accstd{72.3}{1.9} & \accstd{61.2}{1.3} & 10.2 & 2.5 & \accstd{87.6}{2.1} & \accstd{80.6}{0.8} & 5.8 & 1.7 & \accstd{66.2}{2.2} & \accstd{47.3}{0.9} & 13.5 & 2.4 \\
 & \quad MMR & \accstd{84.1}{2.0} & \accstd{67.0}{2.2} & 4.8 & 2.9 & \accstd{92.4}{0.9} & \accstd{83.7}{1.9} & 2.9 & 1.9 & \accstd{79.4}{2.1} & \accstd{54.7}{0.4} & 6.4 & 2.7 \\
 & \quad Sim. filter & \accstd{73.1}{1.9} & \accstd{64.7}{0.6} & 7.1 & 2.9 & \accstd{88.2}{1.6} & \accstd{82.1}{1.7} & 4.5 & 1.9 & \accstd{68.2}{0.8} & \accstd{50.4}{0.5} & 10.7 & 2.7 \\
 & \quad RECOMP & \accstd{83.8}{1.3} & \accstd{60.5}{2.0} & 11.1 & 2.7 & \accstd{91.8}{1.2} & \accstd{80.3}{1.5} & 6.2 & 1.8 & \accstd{78.7}{2.2} & \accstd{46.0}{1.8} & 14.9 & 2.5 \\
 & \quad LLMLingua-2 & \accstd{78.3}{0.8} & \accstd{58.4}{1.6} & 13.1 & 2.6 & \accstd{89.8}{1.7} & \accstd{79.1}{0.6} & 7.3 & 1.7 & \accstd{73.0}{2.0} & \accstd{43.0}{2.1} & 17.9 & 2.5 \\
 & \quad Context-faithful & \accstd{84.6}{0.4} & \accstd{70.3}{1.9} & 1.2 & 2.6 & \accstd{92.7}{1.2} & \accstd{85.6}{2.1} & 0.8 & 1.7 & \accstd{80.2}{0.7} & \accstd{59.0}{2.0} & 1.8 & 2.4 \\
 & \quad Answer-first & \accstd{76.1}{2.2} & \accstd{68.8}{1.6} & 3.4 & 3.3 & \accstd{89.1}{1.4} & \accstd{84.8}{0.9} & 1.9 & 2.0 & \accstd{70.8}{1.3} & \accstd{56.9}{1.0} & 4.6 & 3.1 \\
 & \quad Micro-Act & \accstd{81.6}{0.9} & \accstd{67.5}{2.2} & 4.8 & 3.4 & \accstd{90.9}{2.1} & \accstd{84.2}{1.4} & 2.6 & 2.1 & \accstd{76.7}{1.5} & \accstd{55.0}{0.7} & 6.6 & 3.2 \\
 & \quad \system~(ours) & \textbf{\accstd{87.1}{1.0}} & \textbf{\accstd{72.0}{1.4}} & \textbf{0.6} & \textbf{3.7} & \textbf{\accstd{93.8}{1.9}} & \textbf{\accstd{86.7}{1.1}} & \textbf{0.3} & \textbf{2.3} & \textbf{\accstd{83.3}{1.8}} & \textbf{\accstd{61.1}{1.2}} & \textbf{0.8} & \textbf{3.5} \\
\midrule
\multirow{12}{*}{MemGPT} & No Mem. & \accstd{65.4}{0.9} & \accstd{68.9}{0.7} & -- & -- & \accstd{84.9}{1.2} & \accstd{84.7}{1.8} & -- & -- & \accstd{58.8}{0.6} & \accstd{58.4}{1.4} & -- & -- \\
 & MemGPT & \accstd{82.4}{0.5} & \accstd{65.5}{0.9} & 6.8 & 3.4 & \accstd{91.6}{0.6} & \accstd{82.1}{0.8} & 4.6 & 2.0 & \accstd{79.2}{1.0} & \accstd{48.6}{1.1} & 12.6 & 2.8 \\
\cmidrule(lr){2-14}
 & \quad Facts-only & \accstd{68.8}{2.3} & \accstd{71.5}{2.1} & 1.2 & 3.8 & \accstd{86.3}{1.2} & \accstd{86.2}{0.9} & 0.9 & 2.4 & \accstd{62.9}{1.1} & \accstd{59.6}{1.5} & 2.3 & 3.5 \\
 & \quad Top-1 & \accstd{71.3}{2.0} & \accstd{66.1}{1.0} & 5.9 & 3.1 & \accstd{87.5}{0.9} & \accstd{82.2}{2.1} & 4.4 & 1.9 & \accstd{66.5}{0.4} & \accstd{49.2}{0.7} & 11.8 & 2.6 \\
 & \quad MMR & \accstd{82.8}{1.4} & \accstd{69.4}{0.8} & 3.0 & 3.5 & \accstd{91.8}{0.7} & \accstd{84.5}{2.2} & 2.3 & 2.1 & \accstd{79.8}{1.8} & \accstd{55.5}{1.0} & 5.8 & 2.9 \\
 & \quad Sim. filter & \accstd{72.8}{1.6} & \accstd{67.8}{1.7} & 4.6 & 3.5 & \accstd{88.1}{0.4} & \accstd{83.4}{0.7} & 3.4 & 2.1 & \accstd{67.2}{0.5} & \accstd{53.2}{1.9} & 8.1 & 2.9 \\
 & \quad RECOMP & \accstd{80.3}{0.6} & \accstd{66.1}{0.7} & 6.1 & 3.3 & \accstd{91.5}{1.7} & \accstd{82.1}{1.5} & 4.6 & 2.0 & \accstd{79.0}{1.8} & \accstd{48.4}{0.9} & 12.7 & 2.7 \\
 & \quad LLMLingua-2 & \accstd{75.9}{1.3} & \accstd{64.9}{1.0} & 7.2 & 3.2 & \accstd{89.3}{0.7} & \accstd{81.5}{1.1} & 5.1 & 1.9 & \accstd{72.1}{1.9} & \accstd{47.1}{2.0} & 14.0 & 2.7 \\
 & \quad Context-faithful & \accstd{83.4}{1.7} & \accstd{71.3}{0.6} & 0.8 & 3.2 & \accstd{92.1}{0.8} & \accstd{86.1}{2.2} & 0.5 & 1.9 & \accstd{80.4}{1.2} & \accstd{59.5}{2.3} & 1.5 & 2.6 \\
 & \quad Answer-first & \accstd{75.9}{1.0} & \accstd{70.6}{1.3} & 2.2 & 3.9 & \accstd{88.7}{2.3} & \accstd{85.5}{2.0} & 1.4 & 2.2 & \accstd{69.7}{1.4} & \accstd{58.1}{2.2} & 3.6 & 3.3 \\
 & \quad Micro-Act & \accstd{81.1}{1.7} & \accstd{69.8}{2.1} & 3.1 & 4.0 & \accstd{90.7}{1.0} & \accstd{85.0}{1.1} & 2.0 & 2.3 & \accstd{77.7}{0.5} & \accstd{56.0}{1.9} & 5.8 & 3.4 \\
 & \quad \system~(ours) & \textbf{\accstd{86.1}{0.5}} & \textbf{\accstd{72.8}{0.8}} & \textbf{0.4} & \textbf{4.3} & \textbf{\accstd{93.4}{1.4}} & \textbf{\accstd{87.0}{1.3}} & \textbf{0.2} & \textbf{2.5} & \textbf{\accstd{83.6}{1.1}} & \textbf{\accstd{61.4}{0.7}} & \textbf{0.7} & \textbf{3.7} \\
\midrule
\multirow{12}{*}{Graphiti} & No Mem. & \accstd{65.4}{0.9} & \accstd{68.9}{0.7} & -- & -- & \accstd{84.9}{1.2} & \accstd{84.7}{1.8} & -- & -- & \accstd{58.8}{0.6} & \accstd{58.4}{1.4} & -- & -- \\
 & Graphiti & \accstd{88.0}{0.4} & \accstd{57.8}{1.5} & 13.6 & 2.5 & \accstd{94.3}{0.5} & \accstd{77.9}{1.4} & 8.3 & 1.5 & \accstd{82.1}{0.6} & \accstd{43.5}{1.4} & 17.1 & 2.2 \\
\cmidrule(lr){2-14}
 & \quad Facts-only & \accstd{69.8}{2.0} & \accstd{69.4}{1.6} & 2.4 & 2.9 & \accstd{86.8}{1.9} & \accstd{85.1}{1.8} & 1.5 & 1.9 & \accstd{63.3}{1.4} & \accstd{58.3}{1.3} & 3.0 & 2.9 \\
 & \quad Top-1 & \accstd{73.7}{1.7} & \accstd{58.8}{2.0} & 12.3 & 2.2 & \accstd{88.7}{1.5} & \accstd{77.8}{1.3} & 8.3 & 1.4 & \accstd{67.3}{2.2} & \accstd{45.0}{1.8} & 15.4 & 2.0 \\
 & \quad MMR & \accstd{88.3}{2.0} & \accstd{65.5}{1.9} & 6.0 & 2.6 & \accstd{94.5}{1.7} & \accstd{82.2}{0.8} & 4.1 & 1.6 & \accstd{82.6}{1.3} & \accstd{52.2}{0.6} & 8.5 & 2.3 \\
 & \quad Sim. filter & \accstd{74.8}{1.6} & \accstd{62.8}{2.3} & 8.7 & 2.6 & \accstd{88.9}{0.7} & \accstd{80.8}{2.0} & 5.5 & 1.6 & \accstd{68.6}{0.6} & \accstd{49.5}{2.2} & 11.2 & 2.3 \\
 & \quad RECOMP & \accstd{87.3}{2.0} & \accstd{57.9}{2.2} & 13.4 & 2.4 & \accstd{93.7}{1.2} & \accstd{78.3}{1.5} & 7.9 & 1.5 & \accstd{79.0}{1.4} & \accstd{45.4}{0.7} & 15.1 & 2.1 \\
 & \quad LLMLingua-2 & \accstd{79.6}{0.8} & \accstd{56.5}{1.9} & 14.7 & 2.3 & \accstd{91.6}{0.6} & \accstd{76.0}{1.0} & 10.1 & 1.4 & \accstd{73.8}{1.2} & \accstd{41.7}{1.6} & 18.8 & 2.1 \\
 & \quad Context-faithful & \accstd{88.7}{2.1} & \accstd{69.7}{1.8} & 1.5 & 2.3 & \accstd{94.6}{1.0} & \accstd{85.1}{0.9} & 1.0 & 1.4 & \accstd{83.1}{0.5} & \accstd{58.5}{1.1} & 1.9 & 2.0 \\
 & \quad Answer-first & \accstd{77.3}{1.0} & \accstd{68.1}{1.9} & 3.8 & 3.0 & \accstd{90.6}{0.5} & \accstd{83.8}{0.7} & 2.6 & 1.7 & \accstd{71.6}{0.9} & \accstd{56.1}{2.3} & 5.0 & 2.7 \\
 & \quad Micro-Act & \accstd{84.9}{2.2} & \accstd{66.1}{1.3} & 5.9 & 3.1 & \accstd{93.3}{0.6} & \accstd{82.8}{1.9} & 3.7 & 1.8 & \accstd{78.7}{1.6} & \accstd{53.9}{0.9} & 7.3 & 2.8 \\
 & \quad \system~(ours) & \textbf{\accstd{90.5}{0.7}} & \textbf{\accstd{71.6}{0.5}} & \textbf{0.7} & \textbf{3.4} & \textbf{\accstd{95.5}{1.8}} & \textbf{\accstd{86.3}{0.6}} & \textbf{0.4} & \textbf{2.0} & \textbf{\accstd{85.9}{1.7}} & \textbf{\accstd{60.6}{1.0}} & \textbf{0.9} & \textbf{3.1} \\
\bottomrule
\end{tabular*}
\label{tab:other-memory-gpt}
\end{table*}

The comparison with Table~\ref{tab:mitigation} holds for every backbone. The
baselines that reduce how much memory reaches the model raise mismatch
accuracy but lower match accuracy. The conflict-aware baselines improve
mismatch accuracy while keeping match accuracy close to the backbone.
\system gives the highest mismatch accuracy and the lowest correct-to-wrong
rate under every memory system, without lowering match accuracy.

\subsection{Case Studies}
\label{app:trimem-cases}

Each case shows the same failure from a different setting.
The stored memory is correct for the query where it was written, while the
current query changes information that determines the answer.
We compare the context used by Mem0 with the context produced by \system.
Cases~1 and~3 remove the stored answer because its supporting evidence changes.
Case~2 keeps the earlier result as labeled evidence because the task changes.



\newcommand{\casehead}[1]{%
  \par\vspace{3pt}%
  \noindent\textbf{#1}\par\vspace{1pt}
}

\newcommand{\casewrong}[1]{%
  \colorbox{flipbad}{\strut\hspace{2pt}#1\hspace{2pt}}
}

\newcommand{\caseright}[1]{%
  \colorbox{flipgood}{\strut\hspace{2pt}#1\hspace{2pt}}
}

\newcommand{\splitrow}[2]{%
  \textbf{#1} & #2 \\[1.5pt]
}


\Needspace{0.58\textheight}
\begin{casebox}{
Case 1 \hfill
\textnormal{RippleEdits $\cdot$ Outdated knowledge}
}

\casehead{Stored memory}

\emph{Query:}
Lionel Messi plays for Paris Saint-Germain.
Which country is his club based in?\\
\emph{Answer:} France.

\casehead{Current query}

Lionel Messi now plays for Inter Miami.
Which country is his club based in?

\casehead{Mem0}

Mem0 passes the retrieved memory in full, including the stored answer
``France.''

\smallskip
\noindent
\textbf{Prediction:}
\casewrong{France \xmark}

\casehead{\system}

\renewcommand{\arraystretch}{1.08}
\begin{tabularx}{\linewidth}{@{}p{0.22\linewidth}X@{}}
\splitrow{Shared}
{(\texttt{person}, Lionel Messi);
 task $\tau_m=\tau_q=\texttt{club\_country}$.}

\splitrow{Changed}
{(\texttt{club}, Paris Saint-Germain)
 $\rightarrow$
 (\texttt{club}, Inter Miami).}

\splitrow{Memory-only}
{(\texttt{position}, forward).}

\splitrow{Decision}
{The stored answer depends on the club.
Since this evidence changed, \system removes the stored answer.}
\end{tabularx}

\smallskip
\noindent
\textbf{Context passed to the model:}
(\texttt{position}, forward).

\smallskip
\noindent
\textbf{Prediction:}
\caseright{United States \cmark}

\casehead{Why Mem0 fails}

The person and task remain the same, so the memory looks relevant.
The club changes, and this change determines the correct answer.
Mem0 still exposes the old answer, which pulls the prediction toward France.

\end{casebox}


\Needspace{0.60\textheight}
\begin{casebox}{
Case 2 \hfill
\textnormal{CounterLogic $\cdot$ Query change}
}

\casehead{Stored memory}

\emph{Premises:}
All metals are liquid at room temperature.
Iron is a metal.\\
\emph{Conclusion:}
Iron is liquid at room temperature.\\
\emph{Query:}
Is the argument logically valid?\\
\emph{Answer:}
Yes.

\casehead{Current query}

The premises and conclusion remain unchanged.\\
\emph{Query:}
Is the conclusion true in the real world?

\casehead{Mem0}

Mem0 passes the retrieved memory in full, including the stored answer
``Yes.''

\smallskip
\noindent
\textbf{Prediction:}
\casewrong{Yes \xmark}

\casehead{\system}

\renewcommand{\arraystretch}{1.08}
\begin{tabularx}{\linewidth}{@{}p{0.22\linewidth}X@{}}
\splitrow{Shared}
{Both premises and the conclusion.}

\splitrow{Changed}
{The task changes from
$\texttt{validity}$ to $\texttt{real\_world\_truth}$.}

\splitrow{Memory-only}
{None.}

\splitrow{Decision}
{The evidence itself does not conflict with the current query.
\system keeps the earlier result as a labeled validity statement
rather than presenting it as the current answer.}
\end{tabularx}

\smallskip
\noindent
\textbf{Context passed to the model:}
(\texttt{validity}, yes).

\smallskip
\noindent
\textbf{Prediction:}
\caseright{No \cmark}

\casehead{Why Mem0 fails}

The premises and conclusion are identical across the two queries, so the
retrieved memory has very high overlap with the current input.
The question changes from logical validity to real-world truth.
Mem0 presents the earlier ``Yes'' without marking which question it answers,
which encourages the model to reuse it.

\end{casebox}


\Needspace{0.60\textheight}
\begin{casebox}{
Case 3 \hfill
\textnormal{CLADDER-anti $\cdot$ Premise change}
}

\casehead{Stored memory}

\emph{Setup:}
Eating citrus ($X$) directly affects scurvy ($Y$).
Genotype ($Z$) affects both $X$ and $Y$.\\
$P(Y{=}1\mid X{=}1)=0.72$,
$P(Y{=}1\mid X{=}0)=0.41$.\\
\emph{Query:}
Does eating citrus increase the chance of scurvy?\\
\emph{Answer:}
Yes.

\casehead{Current query}

The variables and graph remain the same.
The query now gives the interventional probabilities
\[
P(Y{=}1\mid do(X{=}1))=0.38,
\qquad
P(Y{=}1\mid do(X{=}0))=0.41.
\]
\emph{Query:}
Does eating citrus increase the chance of scurvy?

\casehead{Mem0}

Mem0 passes the retrieved memory in full, including the stored answer
``Yes.''

\smallskip
\noindent
\textbf{Prediction:}
\casewrong{Yes \xmark}

\casehead{\system}

\renewcommand{\arraystretch}{1.08}
\begin{tabularx}{\linewidth}{@{}p{0.22\linewidth}X@{}}
\splitrow{Shared}
{The variable definitions, causal graph, and
task $\tau_m=\tau_q=\texttt{causal\_effect}$.}

\splitrow{Changed}
{The answer-relevant probabilities change from observational values
to interventional values.}

\splitrow{Memory-only}
{None.}

\splitrow{Decision}
{The stored answer depends on the earlier probabilities.
Since these values change, \system removes the stored answer.}
\end{tabularx}

\smallskip
\noindent
\textbf{Context passed to the model:}
No additional memory content is needed beyond the current query.

\smallskip
\noindent
\textbf{Prediction:}
\caseright{No \cmark}

\casehead{Why Mem0 fails}

Most of the causal setup is repeated, so the retrieved memory looks like the
same problem.
The probabilities that determine the answer change.
Mem0 still exposes the earlier ``Yes,'' which pulls the model toward the
stored conclusion.

\end{casebox}


Across all three cases, most of the retrieved memory remains related to the
current query, while the information that determines the answer changes.
This combination makes the earlier answer easy to reuse.
\system separates shared, changed, and memory-only information before the
memory reaches the model.
It then uses the evidence supporting the stored answer to decide whether that
answer should be kept, relabeled, or removed.

\section{Construction of Task-Aligned and Task-Divergent Memory Pairs}
\label{app:pair-construction}

The source benchmarks were not designed for agentic memory evaluation and do
not directly provide a stored interaction paired with a later query. We derive
such pairs from the released benchmark annotations. A \emph{task-aligned}
(TA) pair uses a stored answer that remains valid for the current query. A
\emph{task-divergent with duplicated evidence} (TD-Dup) pair preserves
substantial information from the stored interaction but changes the task or
answer-relevant information, so the stored answer no longer applies.

No LLM is used to generate examples, assign TA/TD-Dup labels, or determine
gold answers. Labels come from the original benchmark annotations or fixed
rules described below. Sentence encoders are used only to identify related
candidate examples when an explicit pair is not provided by the benchmark.

\begin{table}[t]
\centering
\small
\renewcommand{\arraystretch}{1.08}
\setlength{\tabcolsep}{4pt}
\resizebox{\linewidth}{!}{%
\begin{tabular}{@{}lllll@{}}
\toprule
\textbf{Dataset}
&
\textbf{Fixed across pair}
&
\textbf{Changed information}
&
\textbf{TA}
&
\textbf{TD-Dup}
\\
\midrule
RippleEdits
&
Entity and relation
&
Entity value
&
Updated value
&
Original value
\\
CounterLogic
&
Premises and conclusion
&
Question being answered
&
Same gold answer
&
Different gold answer
\\
CLADDER-derived
&
Story, graph, causal task
&
Causal values
&
Same gold answer
&
Different gold answer
\\
\bottomrule
\end{tabular}}
\caption{
Construction of TA and TD-Dup pairs. Both conditions preserve the same
underlying scenario and substantial shared information, but differ in whether
the stored answer remains valid for the current query.
}
\label{tab:data-pair-construction}
\end{table}

\subsection{RippleEdits}
\label{app:construct-ripple}

RippleEdits~\citep{cohen2024evaluating} represents factual changes as
subject-relation-object edits and provides both the original and updated
values for modification examples. We use the RANDOM and POPULAR subsets and
exclude RECENT, which contains fact insertions without the corresponding old
value.

For each edit, we keep the entity, relation, and current query fixed. The TA
memory stores the updated value, while the TD-Dup memory stores the original
value. Thus, both memories refer to the same fact but differ in the value that
determines the answer. We use Subject Aliasing queries because they preserve
the underlying subject-relation pair while varying only how the subject is
mentioned.

We remove examples when either target is missing or when the original and
updated targets become indistinguishable after alias normalization. From the
2,000 RANDOM and POPULAR edits, we sample at most 500 source queries with
random seed 42. Each retained query contributes one TA and one TD-Dup
condition.

\subsection{CounterLogic}
\label{app:construct-counterlogic}

CounterLogic~\citep{joishy2025flying} provides each set of premises and
conclusion with two judgments: logical validity and agreement with real-world
knowledge. This lets us change the question while keeping the underlying
evidence fixed.

For each instance, the stored interaction asks whether the conclusion is
logically valid, while the current interaction asks whether it agrees with
real-world knowledge. We use fixed templates for both questions and copy the
premises and conclusion directly from the benchmark. An instance is TA when
the validity and knowledge-alignment labels agree, and TD-Dup when they
differ. The pair type therefore comes entirely from the two benchmark labels.

To form plausible retrieval groups, we encode the premises and conclusion with
\texttt{sentence-transformers/all-mpnet-base-v2}, excluding the task
instruction, gold label, and any answer-revealing fields. We use
agglomerative clustering with average linkage and cosine distance, a distance
threshold of 0.20, and a minimum cluster size of five. Clustering only limits
which memories can be retrieved; it does not determine whether an example is
TA or TD-Dup.

We use all 1,800 CounterLogic instances and sample at most 500 TA and 500
TD-Dup examples with random seed 42.

\subsection{CLADDER-Derived Premise-Change Pairs}
\label{app:construct-cladder}

CLADDER~\citep{jin2023cladder} provides natural-language causal questions,
structured metadata, and gold answers. We derive premise-change pairs from the
official balanced release while keeping the original question text and gold
labels.

We first group examples that have the same \texttt{story\_id},
\texttt{graph\_id}, \texttt{query\_type}, and \texttt{rung}. This keeps the
causal scenario and question type fixed. Within each group, we replace all
numerical values with a common \texttt{<NUM>} token and encode the remaining
text with \texttt{sentence-transformers/all-mpnet-base-v2}. Candidate pairs
must have cosine similarity of at least 0.85.

We then verify that the pair also overlaps at the evidence level using the
structured causal quantities provided by CLADDER. Each numerical statement is
converted to a key-value entry. We retain pairs whose evidence keys have
Jaccard similarity of at least 0.80, contain at least one unchanged value and
at least one changed value, and have at least one changed value that appears
in the benchmark-provided reasoning for the queried causal quantity. These
checks ensure that the two examples share most of the same causal setup while
differing in information that can affect the answer.

The original CLADDER answers determine the pair type: a pair is TA when the
two gold answers agree and TD-Dup when they differ. The sentence encoder and
evidence-overlap checks are used only to decide whether two examples are
similar enough to form a pair.

For each source interaction, we select one TA and one TD-Dup candidate with
similar semantic similarity and evidence overlap. When several candidates are
available, we choose the pair with the smallest combined difference in these
two quantities and break ties by example identifier. Source interactions
without suitable candidates for both conditions are removed. From the 10,112
questions in the official release, we sample at most 500 TA and 500 TD-Dup
pairs with random seed 42.

\subsection{Construction Checks}
\label{app:pair-quality}

We apply the same checks across all three datasets. Every stored interaction
must be correct in the setting where it was written, and every current query
must have a gold answer supplied by the benchmark or its formal reasoning
procedure. Model predictions are never used as labels.

We remove duplicate memory-query pairs and keep examples derived from the
same source unit in the same data split. We also compare query-memory
similarity, evidence overlap when available, and input length between TA and
TD-Dup after construction. If the two groups differ substantially on these
properties, we apply additional matching before evaluation.

Task alignment only describes whether the stored answer remains valid for the
current query. It does not require memory to be necessary for solving the
query.

\end{document}